\documentclass{article}

\usepackage{microtype}
\usepackage{graphicx}
\usepackage{subfigure}
\usepackage{booktabs} 

\usepackage[accepted]{icml2023}

\usepackage{amsmath}
\usepackage{bbm}
\usepackage{amssymb}
\usepackage{mathtools}
\usepackage{amsthm}
\usepackage[capitalize,noabbrev]{cleveref}

\theoremstyle{plain}

\theoremstyle{definition}

\theoremstyle{remark}

\usepackage[disable,textsize=tiny]{todonotes}

\newcommand{\quant}{{q_{\Phi}}}
\usepackage{pifont}
\newcommand{\cmark}{\ding{51}}%
\newcommand{\xmark}{\ding{55}}%

\def\nmax{{$n_{\text{max}}$}}

\usepackage{amsmath,amsfonts,bm}

\def\eqref#1{equation~\ref{#1}}

\def\1{\bm{1}}

\def\vc{{\bm{c}}}

\def\vx{{\bm{x}}}

\def\vz{{\bm{z}}}

\def\mX{{\bm{X}}}
\def\mY{{\bm{Y}}}
\def\mZ{{\bm{Z}}}

\DeclareMathAlphabet{\mathsfit}{\encodingdefault}{\sfdefault}{m}{sl}
\SetMathAlphabet{\mathsfit}{bold}{\encodingdefault}{\sfdefault}{bx}{n}

\usepackage{url}

\usepackage[utf8]{inputenc} 
\usepackage[T1]{fontenc}    
\usepackage{booktabs}       
\usepackage{nicefrac}       
\usepackage{microtype}      
\usepackage{wrapfig}

\usepackage{graphicx, multirow}
\usepackage{subfigure}
\renewcommand{\algorithmicrequire}{\textbf{Input:}}
\renewcommand{\algorithmicensure}{\textbf{Output:}}
\usepackage{cleveref}
\usepackage{color, colortbl}
\definecolor{Gray}{gray}{0.9}
\usepackage{arydshln}
\usepackage{enumitem}
\usepackage{threeparttable} 
\usepackage{siunitx}
\usepackage{babel}
\usepackage[font=small]{caption}

\icmltitlerunning{SOM-CPC: Unsupervised Contrastive Learning with Self-Organizing Maps}

\begin{document}

\twocolumn[
\icmltitle{SOM-CPC: Unsupervised Contrastive Learning with Self-Organizing \\ Maps for Structured Representations of High-Rate Time Series}



\icmlsetsymbol{equal}{*}

\begin{icmlauthorlist}
\icmlauthor{Iris A.M. Huijben}{tue,onera}
\icmlauthor{Arthur A. Nijdam}{tue}
\icmlauthor{Sebastiaan Overeem}{tue,kempen}
\icmlauthor{Merel M. van Gilst}{tue,kempen}
\icmlauthor{Ruud J.G. van Sloun}{tue}
\end{icmlauthorlist}

\icmlaffiliation{tue}{Department of Electrical Engineering, Eindhoven University of Technology, The Netherlands}
\icmlaffiliation{onera}{Onera Health, Eindhoven, The Netherlands}
\icmlaffiliation{kempen}{Sleep Medicine Center Kempenhaeghe, Heeze, The Netherlands}

\icmlcorrespondingauthor{Iris Huijben}{i.a.m.huijben@tue.nl}

\icmlkeywords{Contrastive Predictive Coding, Self-Organizing Map, Pattern Recognition, Time Series}

\vskip 0.3in
]



\printAffiliationsAndNotice{}  

\begin{abstract}
  
Continuous monitoring with an ever-increasing number of sensors has become ubiquitous across many application domains. However, acquired time series are typically high-dimensional and difficult to interpret. Expressive deep learning (DL) models have gained popularity for dimensionality reduction, but the resulting latent space often remains difficult to interpret. In this work we propose SOM-CPC, a model that visualizes data in an organized 2D manifold, while preserving higher-dimensional information. We address a largely unexplored and challenging set of scenarios comprising high-rate time series, and show on both synthetic and real-life data (physiological data and audio recordings) that SOM-CPC outperforms strong baselines like DL-based feature extraction, followed by conventional dimensionality reduction techniques, and models that jointly optimize a DL model and a Self-Organizing Map (SOM). SOM-CPC has great potential to acquire a better understanding of latent patterns in high-rate data streams.
\end{abstract}

\section{Introduction}
\label{sec:introduciton}

The improvement and abundance of sensor technology has led to large amounts of high-dimensional, information-rich continuous data streams. However, gaining actionable insights from these data is challenging due to their low interpretability. The main objective of this study is to develop an algorithm for acquiring a structured and interpretable representation of high-rate time series. We define high-rate time series as data streams that are sampled at their Nyquist rate (often hundreds of samples per second), which are distinct from, e.g., (medical) tabular data, which contain sparsely sampled information (e.g. once per hour). We specify an interpretable representation as one that has the ability to be informative and to facilitate exploration of the underlying structure over time \citep{Lipton2018TheInterpretability}. 

%
According to the manifold hypothesis, high-dimensional real-world data lies on a low-dimensional manifold, comprising disentangled latent factors of variation. Dimensionality reduction techniques like Principle Component Analysis (PCA) have conventionally been used to reveal this manifold. However, acquiring an interpretable representation with PCA typically requires omitting many principle components, which may discard important information that can not be linearly projected on the few remaining components. Moreover, additional clustering methods like K-means clustering, are typically needed to find structures or clusters in the the resulting low-dimensional projections.

A Self-Organizing Map (SOM) \citep{kohonen1990self} is an extension of K-means clustering that can be used without the need to first project to the low-dimensional manifold. It creates a low-dimensional interpretable visualization, while still representing the data in multiple dimensions. However, like PCA and K-means, SOMs typically act on features, which need to be selected heuristically and may, therefore, strongly depend on the use case and/or data modality.

Deep learning (DL) models have become popular general data-driven feature extractors, and the sub-area of unsupervised representation learning, is concerned with models that learn to represent data with a set of features, learned without the bias of human annotations. To enhance interpretability, latent space representations of such models are often visualized using a t-distributed stochastic neighbor embedding (t-SNE) \citep{hinton2002stochastic}. Albeit its frequent use, t-SNE does not allow a direct deployment on unseen data as it does not learn a reusable mapping between the multi-dimensional and the low-dimensional space.

DL models have, therefore, also been combined with (joint) clustering objectives in the latent space. Resulting deep-clustering methods are, however, often designed for static data - and if tested on time series - assign a cluster to a full time series rather than to subsequent windows \citep{xie2016unsupervised, Yang2017towards}, preventing exploration of temporally-changing patterns. Also, some require an additional step to create a visually interpretable representation \citep{madiraju2018}, or make use of labels to improve training \citep{Lee2020b}. These labels are by definition unavailable for (yet) hidden patterns.

A sub-area of deep-clustering methods has focused on regularizing autoencoder training with a SOM objective in the latent space \citep{Ferles2018, pesteie2018deep, Fortuin2019,Forest2019DeepMapsv2,Manduchi2021,Forest2021}. However, similar to \citet{mrabah2020adversarial}, we hypothesize that the reconstruction objective of an autoencoder may hamper the clustering or structured representation learning objective: while within-cluster similarities should remain preserved for latent clustering (e.g. preserving only the underlying state of a physiological signal), reconstruction demands a preservation of within-cluster differences as well (e.g. the high-frequency noise). Moreover, in the context of time series representation learning, other self-supervised models - that exploit the temporal dimension - might be more suitable.




Contrastive self-supervised learning approaches have quickly become popular thanks to their superior representation learning performance in many domains (see \citep{le2020contrastive} for a review). While many of these models rely on data augmentations during training in order to construct pairs of similar data points, Contrastive Predictive Coding (CPC) \citep{Oord2019} leverages the temporal dimension for this purpose, making it a natural choice for self-supervised representation learning of time series. In CPC, the temporal dimension not only serves as a pretext task, but simultaneously enforces latent smoothness over time. The contributions of this work are as follows:

\begin{itemize}
    \item We formulate a general framework that captures all previous deep-SOM literature, and inherit it with SOM-CPC, a model for learning structured and interpretable 2D representations of (high-rate) time series by encoding subsequent data windows to a topologically ordered set of quantization vectors. 
    
        
    \item We show that SOM-CPC preserves more information in its 2D representation and exhibits more temporal smoothness in 2D than CPC followed by PCA, a linear classifier or K-means, or directly encoding CPC's latent space to two dimensions.
    
    \item We show that SOM-CPC quantitatively and qualitatively outperforms autoencoder-based deep-SOM models in terms of both clustering and topological ordering, while requiring less auxiliary loss functions and associated hyperparameter tuning.
    
\end{itemize}
 

\section{Preliminaries}
\label{sec:preliminaries}

\subsection{Kohonen Self-Organizing Maps}
\label{sec:soms}

Kohonen's Self-Organizing Map (SOM) \citep{kohonen1990self} is an algorithm to find a visually interpretable topological data representation. It has been found useful to reveal intricate patterns and structure in a plethora of applications \citep{yin2008self}. The algorithm's output, the low-dimensional visualization, is often referred to as a SOM as well.

We define a data point $\vz \in \mathcal{Z}$, and its quantized counterpart
$\quant(\vz) \in \Phi$, with $\Phi=[\phi_1;\ldots;\phi_k]$ a trainable quantization codebook containing $k$ vectors or prototypes $\phi_i \in \mathbb{R}^F, 1\leq i \leq k$. Vector $\quant(\vz) := \Phi[
\operatorname{argmin}_i\big(||\phi_i, \vz||_2^2\big)]$ is thus the `winning quantization vector'. Note that $\Phi$ is updated every training iteration $n$, which we omit for readability.


The codebook vectors are placed on a pre-defined 2D grid of nodes by assigning an xy-coordinate to each vector at initialization. This creates a 2D representation, while each data point $\vz$ lives in $\mathbb{R}^F$, with $F \gg 2$. This is conceptually different than how PCA achieves dimensionality reduction to 2D, where all information in the $3^{\text{rd}}$ and higher principle components is strictly omitted. During SOM training, each $\phi_i$ is updated as follows \citep{kohonen1990self}:
%
\begin{equation}
    \label{eq:som_update_rule}
    \phi_i^{(n+1)} = \phi_i^{(n)} + \eta^{(n)}\mathcal{S}_i\big(\quant(\vz)\big)\big(\vz-\phi_i^{(n)}\big),
\end{equation}
where $\eta^{(n)}$ is an decreasing learning rate. Topological neighborhood structure is promoted via a neighbourhood kernel $\mathcal{S}$ that weighs nodes inversely proportional to their distance with the winning node. A Gaussian kernel is often used, which weighs node $i$ according to:
\begin{align}
    \label{eq:gaussian_kernel}
    &\mathcal{S}_i\big(\quant(\vz)\big) = \exp\Big(-\frac{d^{(n)}_{i}}{2(\sigma^{(n)})^2 }\Big),\hspace{0.5cm} \text{with}\\
    &d_{i}^{(n)} = \big|\big|\mathcal{P}[\quant(\vz)], \mathcal{P}[\phi_i^{(n)}]\big|\big|_2^2   \hspace{1cm} \text{and} \\
    \label{eq:sigma_over_time}
    &\sigma^{(n)} = \sigma^{(0)} \exp(-n/ \lambda), 
\end{align}
where $\mathcal{P}$ projects a codebook vector to its corresponding coordinate on the grid, $\sigma^{(0)}$ denotes the initial standard deviation, and $\lambda$ a decay factor. 
The dependence of $\mathcal{S}_i$ on $d_{i}$, implies a weighing of 1 for the winning node (i.e. distance equals zero), and lower than 1 for neighbour nodes. Other neighbourhood structures have been proposed as well, for example using the four closest neighbours on the grid, resulting in a kernel with a \textit{plus}-shape \citep{Fortuin2019}.

\subsection{Deep-SOM models}
\label{sec:deep_som_models}

Deep-SOM research has focused on combining autoencoders \citep{Ferles2018, pesteie2018deep, Fortuin2019,Forest2019DeepMapsv2,Manduchi2021,Forest2021} with a SOM. These models can broadly be summarized as a vector-quantized variational autoencoder (VQ-VAE) \citep{VandenOordDeepMind}, with a topological organization of the vectors in the quantization codebook: the SOM. The models are trained end-to-end using error backpropagation of both a reconstruction \textit{task} loss $\mathcal{L}_{\text{task}}$ and a loss $\mathcal{L}_{\text{topo}}$ that encourages \textit{topological} ordering in the SOM. In general, a deep-SOM  training objective takes the following form:
%
%
\begin{align}
\label{eq:deep_som_loss}
&\mathcal{L}_{\text{deep-SOM}} = \mathcal{L}_{\text{task}} + \alpha\mathcal{L}_{\text{topo}},\hspace{0.5cm} \text{with}\\
&\mathcal{L}_{\text{topo}}(\vz^{(n)}) = \nonumber \\
\label{eq:topo_loss}
&\hspace{0.5cm} \mathbb{E}_{\mathcal{Z}}\Big[\sum_{i=1}^{k}\mathcal{S}_i\big(\quant(\vz^{(n)})\big)||\vz^{(n)}-\phi_i^{(n)}||^2_2\Big].
\end{align}

Hyperparameter $\alpha$ controls the trade-off. The topological loss replaces the original update rule of the SOM algorithm from \cref{eq:som_update_rule}, and is in practice computed on mini-batches of data to approximate the expectation. The features in $\mathcal{Z}$ are jointly optimized via the parameters of the encoder, and thus also depend on $n$ now. To prevent clutter we will, however, omit the (n)-superscript in the following.




\citet{Fortuin2019} propose SOM-VAE. As opposed to VQ-VAE, SOM-VAE has two decoders, as it also decodes the continuous latents.
The topological loss was split in a \textit{commitment} loss (committing the winning codebook vector to $\vz$ and \textit{vice versa}) and a \textit{SOM} loss (pulling the codebook vectors of the neighbours to $\vz$): \mbox{$\mathcal{L}_{\text{topo}} = \mathcal{L}_{\text{commitment}} + \frac{\beta}{\alpha} \mathcal{L}_{\text{SOM}}.$} Formally:
%
%
\begin{align}
\label{eq:commitment_loss}
&\mathcal{L}_{\text{commitment}} = \mathbb{E}_{\mathcal{Z}}\Big[||\vz-\quant(\vz)||^2_2\Big], \hspace{1cm} \text{and}\\
\label{eq:SOM_loss}
&\mathcal{L}_{\text{SOM}} = \mathbb{E}_{\mathcal{Z}}\Big[\sum^{k}_{\substack{i=1;\\ \phi_i\neq \quant(\vz)}}\mathcal{S}_i\big(\quant(\vz)\big)||\operatorname{sg}[\vz]-\phi_i||^2_2\Big],
\end{align}
with $\operatorname{sg}[\cdot]$ a gradient blocker, making the encoder parameters independent of the quantization error of the neighbour nodes. The authors chose a plus-shaped neighbourhood kernel $\mathcal{S}$. Note that for $0 < \beta/\alpha < 1$, this plus-kernel is a coarse approximation of the Gaussian kernel. The sum of the reconstruction losses $\mathcal{L}_{\text{recon,cont}}$ and $\mathcal{L}_{\text{recon,disc}}$ from the continuous and discrete decoder, respectively, yield the total task loss $\mathcal{L}_{\text{task}}$, which is combined with the topological loss to create the training objective of the SOM-VAE model. 


SOM-VAE-prob \citep{Fortuin2019} and (T)-DPSOM \citep{Manduchi2021} are extensions of SOM-VAE. SOM-VAE-prob enforces smoothness over time by adding a transition loss (multiplied by $\gamma$) to optimize a first-order Markov model to learn the node transition probabilities, and a smoothness loss (multiplied by $\tau$ in their work, we will use $\zeta$) to minimize the quantization error of highly probable transitions. DPSOM is a probabilistic model, based on a VAE with a non-degenerate approximate posterior \citep{kingma2013auto} with soft cluster assignment and a cluster assignment hardening (CAH) loss \citep{xie2016unsupervised}. T-DPSOM additionally incorporates a temporal smoothness loss, and an LSTM, which aims to predict the future latent space. 
%
%
Note that the measures for additional temporal smoothness in SOM-VAE-prob and T-DPSOM impose additional loss-weighing hyperparameters that need tuning, possibly for each data modality separately. This smoothness is already naturally embedded in the task objective of our model (see \cref{sec:cpc}). The probabilistic additions in the (T-)DPSOM model, with respect to SOM-VAE, are orthogonal to the developments in this work.

\begin{figure*}
    \centering
    \includegraphics[width=0.9\linewidth,trim={4.6cm 7.5cm 4.3cm 5.5cm},clip]{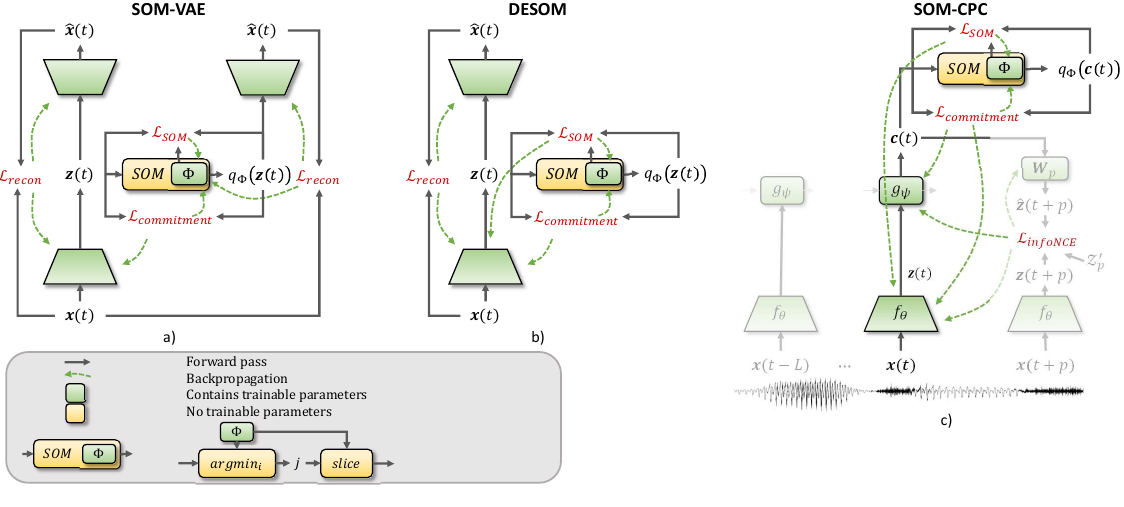}
    \\[-1em]
    \caption{Architectures of different deep-SOM models including the gradient paths in green. a) SOM-VAE \citep{Fortuin2019} b) DESOM \citep{Forest2021} c) SOM-CPC (ours). The two decoders in the SOM-VAE model are independent and have their own trainable parameters, while the visualized encoders in the SOM-CPC model are all the same (i.e. parameters are shared). The $g_{\psi}$ block in the SOM-CPC model indicates an autoregressive component (e.g. a GRU), and $\mathcal{Z}'_p$ refers to a set of drawn negative embeddings.
    \label{fig:architectures}
    }
    \vspace{-0.4cm}
\end{figure*}


\citet{Forest2021} propose Deep embedded SOM (DESOM). Compared to SOM-VAE, the decoder on the discrete space is omitted (therewith also $\mathcal{L}_{\text{recon,disc}}$), gradients from $\mathcal{L}_{\text{SOM}}$ to the encoder are not being blocked (i.e. $\operatorname{sg}[\cdot]$ is removed from \cref{eq:SOM_loss}), and the topological loss $\mathcal{L}_{\text{topo}}$ is given by \cref{eq:gaussian_kernel}, i.e. with a Gaussian neighbourhood function with decaying variance.
In a short work, \citet{Forest2019DeepMapsv2} speculate about adding an LSTM in the latent space to train a SOM on sequential data, and refer to this model as LSTM-DESOM. Figure \ref{fig:architectures}a-b visualizes the SOM-VAE and DESOM architecture.

\section{SOM-CPC}
\label{sec:methodology}

\subsection{Motivation}
Discovering new patterns in high-dimensional data using SOMs requires features that are both suitable for SOM organization \textit{and} accurately reflect the data. A desirable property of a feature extractor is that it inverts the data-generating process, inferring a latent low-dimensional state that carries useful information. This data-generating process is in general unknown, implicit and highly non-linear. Feature learning can thus be formulated as a non-linear independent component analysis problem, which was proven to be non-identifiable \citep{hyvarinen1999nonlinear}. However, recent advances showed that the problem becomes identifiable under the assumed presence of an auxiliary variable \citep{Hyvarinen2018}. Such an auxiliary variable (e.g. a temporal component) is not present in plain autoencoders, but the contrastive learning paradigm was shown to conform to this assumption \citep{Hyvarinen2018, Zimmermann2021}. 

In line with that, \citet{Oord2019} showed that minimizing the InfoNCE loss in the contrastive learning framework CPC maximizes the mutual information between a context vector that summarizes past and current information, and a latent space from a future data window. CPC thus results in features that encode a slowly-changing underlying state. Encoding (high-frequency) information that is less-shared across windows is thus not needed to minimize InfoNCE, while such within-cluster/state differences typically do need to be encoded for reconstruction. This is hypothesized to make autoencoders less suitable to do feature extraction for clustering purposes \citep{mrabah2020adversarial}.

We thus propose to leverage CPC as a feature extractor, and jointly optimize it with a SOM, which results in SOM-CPC: a representation learning model that learns to map windows of time series data to a structured 2D grid for the purpose of pattern discovery.

\subsection{Algorithmic Details}
\label{sec:cpc}
We introduce \mbox{$\mX = [\ldots,\vx(t), \vx(t+1),\ldots]$}, a sequence of non-overlapping data windows \mbox{$\vx(t) \in \mathbb{R}^{ch \times T}$}, with $ch$ the number of channels, $T$ the number of samples in a window, and $t$ the index of the window. For brevity we omit this window index when possible. Following \citet{Oord2019}, an encoder, parameterized by $\theta$, maps each window $\vx$ to a latent representation $\vz = f_{\theta}(\vx) \in \mathbb{R}^{F}$, with $F$ the number of features.  $\mZ$ includes the embeddings of all windows in $\mX$. The sets containing all available data streams $\mX$ and corresponding embeddings $\mZ$ are denoted with $\mathcal{X}$ and $\mathcal{Z}$, respectively. A causal auto-regressive (AR) module $g_{\psi}$ parameterized by $\psi$, e.g. a gated-recurrent unit (GRU), subsequently aggregates the current and $L$ previous embeddings into a context vector $\vc(t) \in \mathbb{R}^{F}$. From this context, predictions are made for $P$ future (or `positive') embeddings $\vz(t+p)$, with $p \in \{1, \ldots, P\}$. The task objective in SOM-CPC maximizes the match (expressed as a dot-product) between these predictions and the positive embeddings, as compared to this match for $N$ `negative' embeddings. These negatives may be sampled across the dataset (i.e. from $\mathcal{Z}$), or within the same time-series (i.e. from $\mZ$). The task objective, being the InfoNCE loss \citep{Oord2019}, is defined as:
\begin{align}
    \label{eq:CPC_loss}
    &\mathcal{L}_{\text{task}} := \mathcal{L}_{\text{InfoNCE}} = \frac{1}{P} \sum_{p=1}^P \mathcal{L}_p, \hspace{0.3cm}\text{with}\hspace{0.3cm} \mathcal{L}_p = \\
    \label{eq:CPC_loss_per_p}
    &- \underset{\mathcal{X}}{\mathbb{E}} \Big[\log \frac{\exp\Big(\vz(t+p)\mathbf{W}_p \vc(t)\Big)}{\sum_{\vz' \in \mathcal{Z}_p'\cup \{\vz(t+p)\}} \exp\Big(\vz'\mathbf{W}_p \vc(t)\Big)} \Big] ,
\end{align}
\noindent with $\mathcal{Z}_p' \subset \mathcal{Z}$ the embeddings of drawn negative samples ($|\mathcal{Z}_p'| = N$), and $\mathbf{W}_p \in \mathbb{R}^{F \times F}$ a trainable linear predictor between the current context vector and the $p^{\text{th}}$ future embedding. 

The context vector is not only used to predict future embeddings, it is also the input to the SOM module that selects the winning node.  We use a use a 2D (square) grid for the SOM nodes to acquire visual interpretability. The SOM is optimized using the topological loss $\mathcal{L}_{\text{topo}}$, as defined in \cref{eq:topo_loss}, with a Gaussian neighbourhood kernel $\mathcal{S}$ (see \cref{eq:gaussian_kernel}).
Depending on the use case, it was found to not always be necessary, or even beneficial (due to higher risk of overfitting), to use an AR module $g_{\psi}$ to aggregate causal context into the current embedding. If the AR module is not used, the future predictions are made directly from the current (continuous) latent space $\vz(t)$ instead of $\vc(t)$. Likewise, $\vz(t)$ rather than $\vc(t)$ is being quantized by the $\operatorname{SOM}$ module. Depending on the presence of this AR module, both $\mathcal{L}_{\text{topo}}$ and $\mathcal{L}_{\text{task}}$ are thus computed on either $\vz(t)$ or $\vc(t)$.

All model elements are jointly optimized, with the training objective:
$\mathcal{L}_{\text{SOM-CPC}} = \mathcal{L}_{\text{task}} + \alpha\mathcal{L}_{\text{topo}},$
which adheres to the general objective of a deep-SOM model as formulated in \cref{eq:deep_som_loss}. Figure \ref{fig:architectures}c visualizes SOM-CPC, and its gradient paths in green. The initial standard deviation of the Gaussian neigbhourhood kernel was set to $\sigma^{(0)}=\sqrt{k}/2$, ensuring that the full grid is part of the neighbourhood at the start of training. 
Pseudo-code can be found in \cref{app:general_details} (Algorithm \ref{algorithm}) and code is published on \url{https://github.com/IamHuijben/SOM-CPC.git}.


\subsection{Performance Evaluation}
\label{sec:performance_evaluation}

\citet{forest2020survey} provide a taxology of SOM metrics that distinguishes \textit{external} vs \textit{internal} and \textit{topological} vs \textit{clustering} metrics. External metrics are related to labels (which are not used during unsupervised training), while internal metrics do not depend on such information. Topological metrics assess the topological ordering (i.e. neighbourhood relations) of the SOM, while clustering metrics are more related to, for example, pureness of nodes.

To evaluate clustering performance, linked to external labels, we leverage purity and the normalized mutual information (NMI). The latter corrects for a high number of clusters (i.e. nodes), which could easily lead to high pureness, but leaves the NMI more conservative. 

Even though scoring high on external metrics is not the main goal of a representation learning model like SOM-CPC, we do report them as it provides an indication of how well information was preserved. To compute regression/classification performance, we first `color' (or label) each node with the most occurring (for discrete labels) or median (for continuous labels) label from the training set. The test set predictions are then converted from node indices to label predictions by using these colorings. Regression performance is expressed as the average squared regression error with the target: $\operatorname{SE}_{\text{target}}$. Classification performance is reported with Cohen's kappa \citep{Cohen1960}, a commonly used metric that corrects for correctness by chance.

Topographic performance is measured using the (internal) Topographic Error (TE) \citep{kiviluoto1996topology}, which reports the fraction of windows (between 0 and 1) for which the winning and second-best winning node are not neighbours in the SOM (lower is better). Finally, to measure whether a time series conveys a smooth trajectory through SOM space, we measure the average Euclidean distance (denoted $\ell_{2,\text{smooth}}$) between all subsequent windows in each time series. The lower this value, the less frequently large jumps in the 2D map occur. Note that in extreme cases where many windows collapsed to the same node, both the TE and the average $\ell_{2,\text{smooth}}$ metric are artificially pushed down. We can thus only interpret these metrics in conjunction with earlier-mentioned clustering and classification metrics. Mathematical definitions of all metrics are provided in \cref{app:general_details}.

 
\section{Experiments}
\label{sec:experiments}

\begin{table*}[t]
\caption{Mean and one std. dev. across all test set series of synthetic data. Results for varying values of $\alpha$, $\gamma$, and $\zeta$ can be found in \cref{tab:toy_results_extended}, \cref{app:toy_extended_res}. SOM-CPC clearly outperforms all baselines. Models below the dashed line are ablations. Regression plots and SOMs of models with a * are visualized in \cref{fig:som_maps_toy}a-c. }
\label{tab:toy_results}
\centering
   \begin{tabular}{p{0.1cm}lccc|lll}
    	\hline
    	\rowcolor{Gray}
    	& \textbf{Model} & $\boldsymbol{\alpha}$ & $\mathcal{S}$ & $\mathcal{L}_{\text{SOM}} \operatorname{sg}[\cdot]$ &  $\operatorname{SE}_{\text{target}}$ & \textbf{$\ell_{2,\text{smooth}}$}  & \textbf{TE} \\\hline\hline
    	& CPC + linear classifier & - & - & - & ~~2.62$\pm$~~2.37 & - & -
    	\\
    	& CPC + K-means & - & - & - & ~~1.09$\pm$~~~~.62  & - & - 
    	\\\hline
    	& CPC ($F=2$)~+ linear classifier & - & - & - & 25.01$\pm$42.94
        & - & - 
        \\
        & CPC ($F=2$)~+ K-means & - & - & - & ~~~~.76$\pm$~~1.31
        & - & - 
    	\\
    	& CPC + PCA + linear classifier & - & - & - & 42.81$\pm$58.12  & - & - 
    	\\
    	& CPC + PCA + K-means & - & - & - & ~~4.42$\pm$~~9.01  & - & - 
    	\\\hline
        * & SOM-VAE & .1   & Plus &  \cmark  & ~~8.02$\pm$~~4.58	&	2.41$\pm$.68	&	.28$\pm$.07
        \\
        & SOM-VAE & 1e-3 & Gaussian & \cmark  & 11.60$\pm$25.43	&	1.92$\pm$.34	&	.06$\pm$.02
        \\ 
        & SOM-VAE-prob  & .1 & Plus & \cmark  & 20.10$\pm$48.80	&	3.15$\pm$.73	&	.63$\pm$.05 
        \\
         * & DESOM & .1 & Gaussian & \xmark & 10.77$\pm$10.85	&	2.20$\pm$.44	&	.06$\pm$.02
        \\\hline
         * & SOM-CPC (ours) & 
          1e-4 & Gaussian & \xmark & ~~~~.72$\pm$~~1.08	&	1.37$\pm$.37	&	\textbf{.02$\pm$.01 }\\\hdashline
        \hspace{-0.5cm}\parbox[t]{2mm}{\multirow{4}{*}{\rotatebox[origin=c]{90}{\underline{~Ablations~}} }} & SOM-CPC & 1e-2 & Gaussian & \cmark & \textbf{~~~~.47$\pm$~~~~.48}	&	\textbf{~~.99$\pm$.24}	&	.07$\pm$.04
\\
   	    & SOM-CPC & 1e-4 & Plus & \xmark & ~~1.71$\pm$~~1.15	&	2.46$\pm$.51	&	.30$\pm$.06
\\
    	& SOM-CPC & 1e-2 & Plus & \cmark  & ~~1.16$\pm$~~~~.61	&	1.85$\pm$.26	&	.12$\pm$.04
\\
    	& CPC + SOM (disjoint) & - & Gaussian & - & ~~~~.84$\pm$~~1.14 & 1.47$\pm$.50 & .03$\pm$.01
        \\\hline
    \end{tabular}
\end{table*}

SOM-CPC is compared to several 2D representation learning methods. First, deep-SOM models with a reconstruction task loss (i.e. SOM-VAE(-prob), and \mbox{$\text{(GRU-)DESOM)}$}. Second, vanilla CPC with a high-dimensional latent space (\mbox{$F \gg 2$}) \citep{Oord2019}, followed by PCA for dimensionality reduction to 2D. Third, CPC with a 2D latent space (\mbox{$F=2$}). For the CPC-based models, linear and non-linear read-out is, respectively, tested using a linear neural classifier, and K-means clustering with the same number of clusters as the number of nodes used in SOM-CPC. High-dimensional vanilla CPC (\mbox{$F \gg 2$}) without additional dimensionality reduction is, moreover, tested as it sets an upper bound for the amount of information that can be preserved given the encoder architecture, while not providing an interpretable 2D representation. The same encoder architecture is used for all models that are compared in a single application domain, and all models are run with the same seed for randomization. We benchmarked our implementation of SOM-VAE against reported results by \citet{Fortuin2019} (see \cref{app:benchmarks}). Details on model architectures and training settings for the different applications can be found in \cref{app:toy_architecture}, \ref{app:sleep_architectures}, and \ref{app:audio_architectures}.

\begin{figure*}
    \centering \includegraphics[scale=0.4,trim={0cm 7cm 10cm 0cm},clip,page=1]{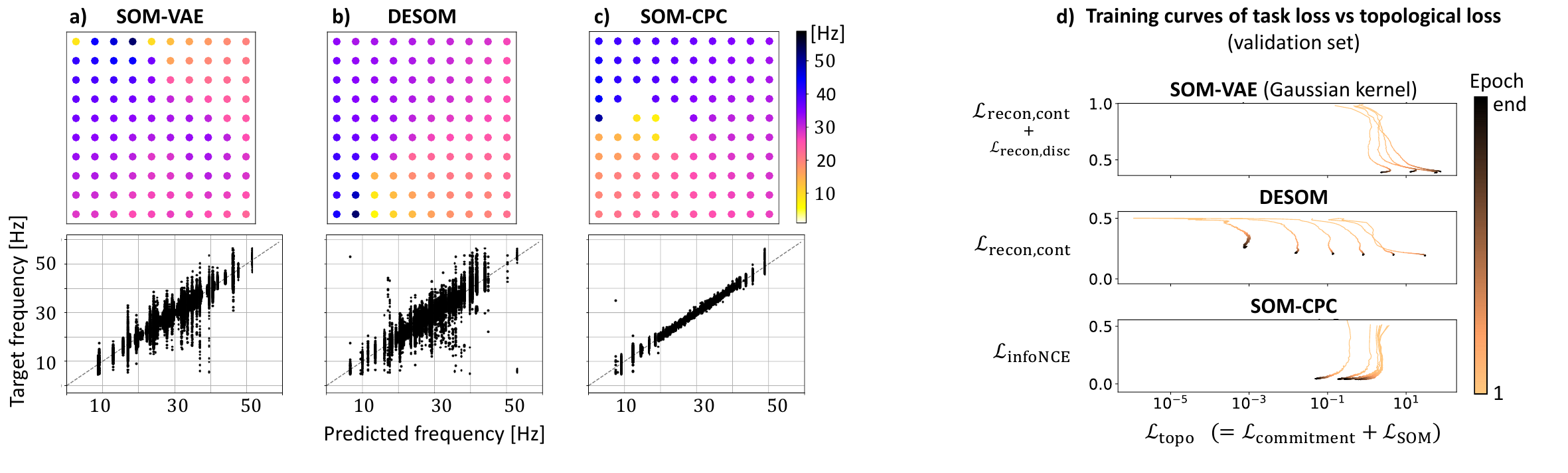}
    \caption{SOMs and regression plots for SOM-VAE (a), DESOM (b) and SOM-CPC (c). Both DESOM and
SOM-CPC show a gradual change of frequency over the grid, but the regression error $\operatorname{SE}_{\text{target}}$ is lower for SOM-CPC, which can also be
seen from the regression plot, where the predicted window frequencies are plotted against the target frequencies (i.e. training set median label) for the node on
which the window was mapped. d) Task loss versus the topological loss for SOM-VAE (with Gaussian neighbourhood), DESOM and SOM-CPC (both with and without $\operatorname{sg}[\cdot]$). The different curves display various values of $\alpha$, for which the DESOM model seems most sensitive. The SOM-CPC models follow a smooth optimization curve, minimizing both the task and topological loss, while these losses seem to be more conflicting in SOM-VAE and DESOM training.}
    \label{fig:som_maps_toy}
\end{figure*}

\begin{figure}[t]
    \centering
    \includegraphics[width=\linewidth,trim={0cm 10cm 26.5cm 0cm},clip]{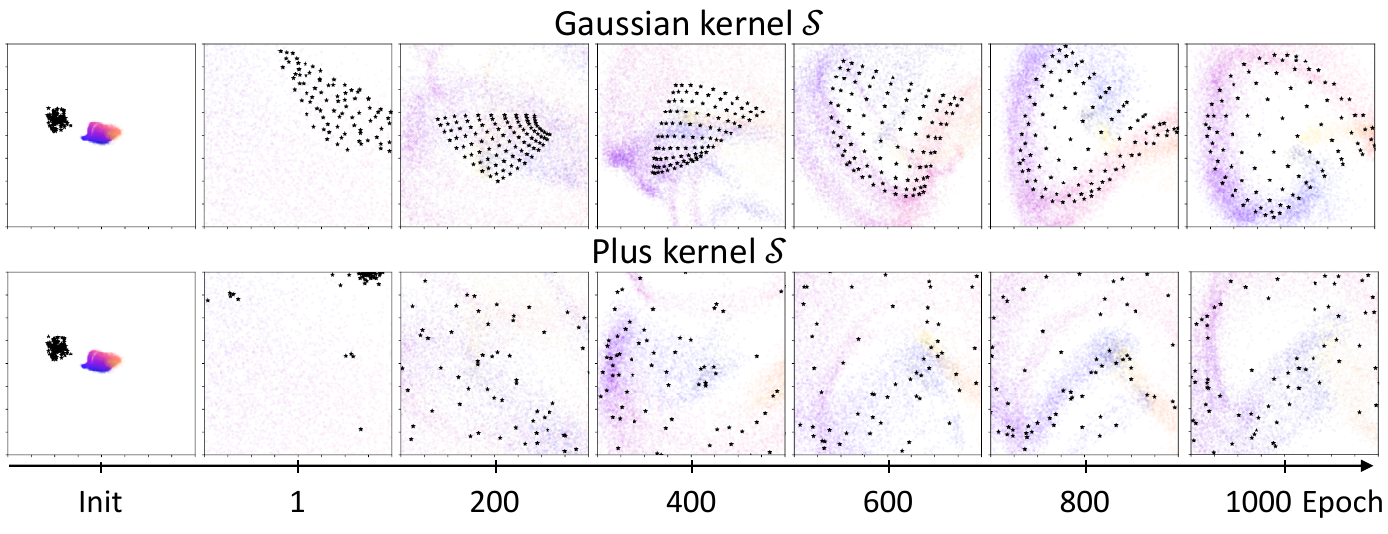}
    \\[-0.5em]
    \caption{Progression of SOM training (nodes in black) in the SOM-CPC model, using either a Gaussian (top) or plus neighbourhood (bottom) kernel, with the PCA projection of the test set latent space plotted behind the nodes. The Gaussian kernel enforces a more strict organization of SOM nodes that quantize the latent space by placing more nodes at higher density areas.}    
    \label{fig:SOM_over_time}
    \vspace{-0.4cm}
\end{figure}

\subsection{Synthetic Data
\label{sec:synthetic_data}}

\textbf{Data generation:} A synthetic dataset was created, consisting of sinusoids with an initial frequency sampled from a uniform distribution between 20 and 40 Hz. The frequency of the signals was altered over time according to a random walk process with a step size of $0.1$~Hz. 
As such, at each time step (i.e. sample), the signal's frequency either increased with 0.1 Hz (with probability $p_{\text{up}}=0.1$), decreased with 0.1 Hz ($p_{\text{down}}=0.1$), or remained constant ($p_{\text{c}}=0.8$). 
In case the random walk crossed either 1 or 60 Hz, the probabilities were (temporarily) altered to
[$p_{\text{up}}$, $p_{\text{c}}$, $p_{\text{down}}$] = [$0.5, 0.5, 0$] or [$p_{\text{up}}$, $p_{\text{c}}$, $p_{\text{down}}$] = [$0.0, 0.5, 0.5$], respectively. 
All series were finally corrupted with an additive white Gaussian noise vector $\mathbf{\epsilon} \sim ~\mathcal{N}(0,0.01)$. Formally, each generated signal took the form:
$\vx[n] = \sin\Big(2\pi \frac{f[n-1] + \Delta f}{f_s}n \Big) + \mathbf{\epsilon},$
with $f[n=0] \sim U[20,40]$, \mbox{$\Delta f \sim \operatorname{Categorical}([p_{\text{up}}, p_{\text{c}}, p_{\text{down}}])$}, and $f_s = 128$~Hz the sampling frequency.
A total of 200 of such time series, each of 5 minutes, were generated, and labels were defined per 1-second window by taking the median frequency. The set was randomly divided into a training ($n=100$), validation ($n=50$), and test split ($n=50$).

\textbf{Comparison to deep-SOM baselines:} 
Table \ref{tab:toy_results} shows that SOM-CPC outperforms all deep-SOM baselines on all metrics. Moreover, \cref{fig:PCA_toy} in \cref{app:toy_extended_res} reveals that the latent space of SOM-CPC, when projected on its two principle components, shows a better organization of the signal frequency than the projections of the SOM-VAE and DESOM models. Figure \ref{fig:som_maps_toy}a-c display the resulting SOMs (colored with the median test set labels) for the same three models. Uncolored nodes in the SOM were not assigned in the test set. Interestingly, although the SOM for the DESOM and SOM-CPC model look similar, the $\operatorname{SE}_{\text{target}}$ is higher for the DESOM model, which can also be seen from the regression plots below the SOMs. 

The addition of two temporal losses in the SOM-VAE-prob model, as compared to SOM-VAE, did deteriorate the given metrics, even though a range of values for multipliers $\alpha$, $\gamma$ and $\zeta$ was tested (see \cref{tab:toy_results_extended} in \cref{app:toy_extended_res} for the full sweep). The deterioration of the results can be explained by the difficulty of finding the correct scaling factors for these additional losses. Note that the SOM-CPC model automatically incorporates smoothness over time thanks to the nature of the CPC task loss, therewith preventing additional hyperparameter tuning. 

Additionally, we study the optimization behaviour of SOM-VAE, DESOM, and SOM-CPC by plotting the progression of the task versus the topological loss during training (see \cref{fig:som_maps_toy}d). To make a fair comparison, we plot the SOM-VAE models that are trained with a Gaussian neighbourhood kernel. Different curves in the graphs indicate runs with varying values for $\alpha$, and the line color's gradient denotes the training iteration. The SOM-CPC graphs include the models run with and without gradient detachment of $\mathcal{L}_{\text{SOM}}$ to the encoder. It can be seen that both losses jointly minimize in SOM-CPC training, while there is a counteracting effect visible for SOM-VAE, and a high influence of the value of $\alpha$ for DESOM training.

\textbf{Comparison to other  baselines:} 
CPC (with $F=2$), and CPC followed by PCA, resulted in a much higher regression error $\operatorname{SE}_{\text{target}}$ than SOM-CPC when using linear read-out (see \cref{tab:toy_results}). Non-linear K-means clustering improved performance for both cases, but only for CPC with $F=2$, performance nearly reached SOM-CPC performance. Later we will see that optimizing CPC with $F=2$ can hamper optimization for more intricate data spaces (see \cref{sec:audio}). Interestingly, regression performance of SOM-CPC was found to be even slightly better than that of the vanilla multi-dimensional CPC model (with $F=128$), both for linear classification and K-means. This could be explained by the additional regularization that the SOM provides in SOM-CPC training. 

\textbf{Ablations:}
We perform several ablation experiments on SOM-CPC, which are reported below the dashed line in \cref{tab:toy_results}. Blocking the gradients of the neighbour nodes with respect to the encoder during training ($\mathcal{L}_{\text{SOM}} \operatorname{sg}[\cdot]$ column) slightly improved the regression error and temporal smoothness, but decreased the topographic error. Looking at models run with various values for $\alpha$ (reported in \cref{tab:toy_results_extended}, \cref{app:toy_extended_res}), the effect of gradient blocking can be considered small and ambiguous when considering different metrics. Disjoint training (CPC + SOM) resulted in a less smooth trajectory over time through the 2D SOM space, seen from the higher $\ell_{2,\text{smooth}}$. Using a plus neighbourhood kernel instead of a Gaussian kernel decreased performance on all three metrics. The increase in TE, is well explainable by the fact that a plus kernel takes into account fewer neighbours (at least at the start of training) and therefore has more difficulty to find a good topological mapping. \Cref{fig:SOM_over_time} shows the development of the SOM node spread (projected on top of a PCA projection of the continuous test set latents), during training for SOM-CPC with the two type of kernels. It can indeed be seen that the Gaussian kernel enforces a more strict topological organization. Interestingly, the kernel does not only influence the codebook vectors, but also seems to influence the organization of the latent space, seen from the differently-shaped PCA projections in the background.

\subsection{Physiological Data}
\label{sec:modelling_sleep}

\textbf{Data:} We analyse SOM-CPC on physiological data. To this end, we selected whole-night polysomnography recordings which comprise a broad range of widely-used physiological sensors: electroencephalography (EEG), electromyography (EMG), and electrooculography (EOG). We used subset 3 of the Montreal Archive of Sleep Studies (MASS) database \citep{OReilly2014}\footnote{\url{http://ceams-carsm.ca/mass/}}, consisting of polysomnography recordings for which every 30-second window is labelled with a sleep stage label from $\{\text{N1, N2, N3, REM, Wake}\}$. The 62 recordings (from 62 unique subjects) were randomly split into a training ($n=48$), validation ($n=8$) and hold-out test set ($n=7$). Details on the data preprocessing can be found in \cref{app:sleep_data_preprocessing}.


\begin{figure}
\centering
\includegraphics[width=\linewidth,trim={0cm 8cm 12.5cm 0.cm},clip]{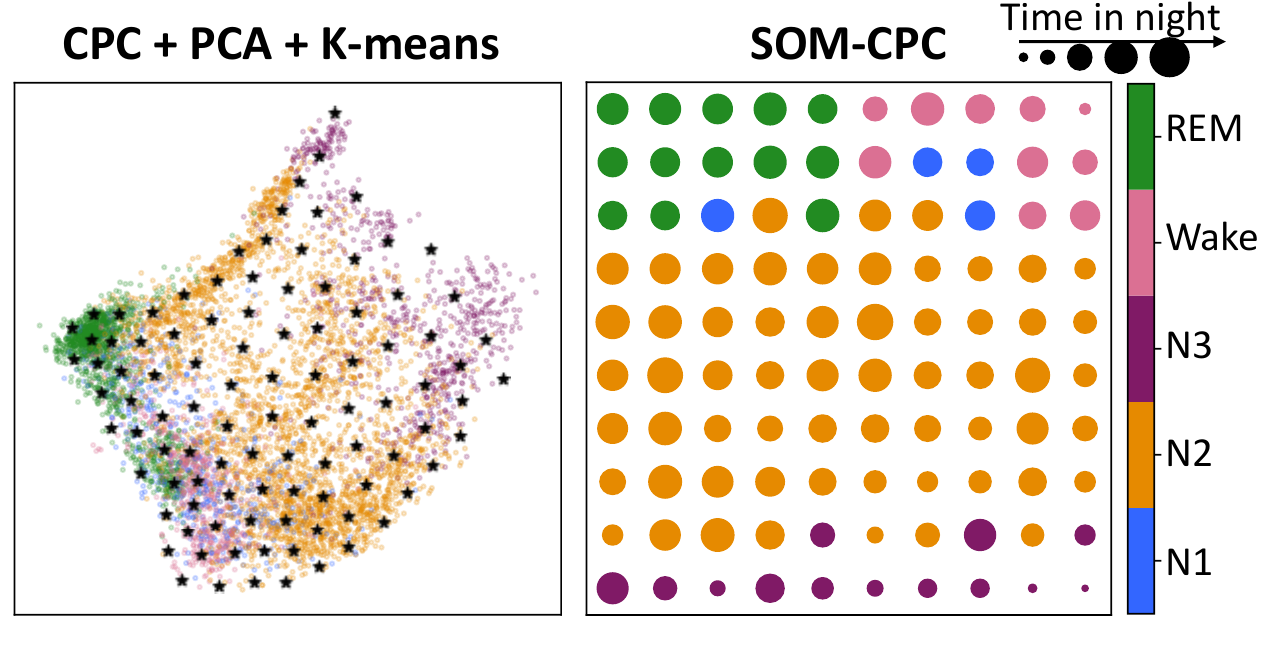}
\caption{Deep sleep N3 is isolated from light sleep N1, Wake and REM sleep with a cluster of medium-deep sleep N2. \label{fig:sleep_som_maps}}
\vspace{-0.4cm}
\end{figure}

\textbf{Comparison to deep-SOM baselines:}
\Cref{tab:sleep_results} in  \cref{app:extended_sleep_results} shows that SOM-CPC again clearly outperformed SOM-VAE and DESOM on all metrics. Using SimCLR \citep{Chen2020} - a popular contrastive learning framework - as task objective instead of CPC's InfoNCE loss also deteriorated performance (see \Cref{app:extended_sleep_results}). 

\textbf{Ablations:} Whether or not the gradients of the SOM loss were stopped towards the encoder did not greatly influence SOM-CPC performance. Topological ordering, measured by TE, and temporal smoothness ($\ell_{2,\text{smooth}}$) deteriorated when changing the Gaussian kernel to a plus kernel, or training CPC and SOM disjointly \citep{huijben2022}.

\textbf{Comparison to other baselines:}
SOM-CPC's classification performance was higher than that of CPC with \mbox{$F=2$}, and CPC followed by PCA. \Cref{fig:sleep_som_maps} shows the test set PCA projection of the latent space of CPC ($F = 128$), with the K-means nodes as black stars (left), and the SOM (right) trained by the SOM-CPC model (nodes are colored with the most-occurring label in the test set). Both visualizations show similar clustering patterns: deep sleep N3 is isolated from lighter forms of sleep (i.e. N1, Wake and REM sleep) by a thick cluster of medium-deep sleep N2. However, the higher performance of SOM-CPC (see \cref{tab:sleep_results}) indicates that more information is preserved in the 2D space resulting from the SOM-CPC model. The size of the nodes in the SOM of SOM-CPC indicates the average time in the night of windows on that node. A difference is visible in node sizes within the Wake, N2 and N3 clusters, suggesting a possible existence of different sub-categories of sleep within the pre-defined sleep stages.

\subsection{Audio}
\label{sec:audio}

\textbf{Data:} We use a subset of the publicly available LibriSpeech dataset \citep{panayotov2015librispeech}\footnote{\url{https://www.openslr.org/12}}, which contains multiple minute-long English voice recordings of 251 different speakers, sampled at 16~KHz. We leveraged the publicly available train-test split, as provided by \citet{Oord2019}\footnote{\url{https://drive.google.com/drive/folders/1BhJ2umKH3whguxMwifaKtSra0TgAbtfb}}, and created an additional validation set by randomly selecting  $25\%$ of the training set. 
Recordings of the ten speakers with the longest recording time were selected to alleviate computational burden. 
This resulted in a total of 150.9, 54.6, and 46.5 minutes in the training, validation, respectively test set. Model and training details can be found in \cref{app:audio_architectures}.



\begin{figure}
\centering
\includegraphics[width=\linewidth,trim={0cm 10cm 9.3cm 2.2cm},clip,page=1]{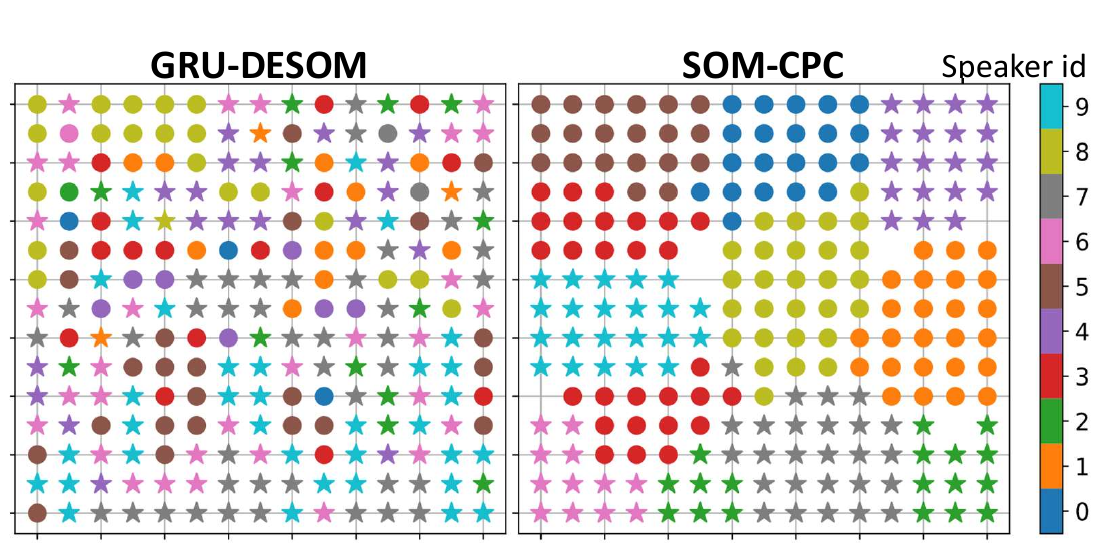}
\caption{SOM-CPC is able to better cluster different speakers than GRU-DESOM. Stars denote women and dots are men.}
\label{fig:audio_som_maps}
\vspace{-0.4cm}
\end{figure}

\textbf{Comparison to deep-SOM baselines:}
Table \ref{tab:audio_results} in \cref{app:extended_audio_results} shows the results of SOM-CPC (which includes a GRU for this dataset), compared to different variants of the DESOM model. SOM-CPC outperforms all DESOM variants by a wide margin and for all choices of the $\alpha$ parameter. The difference in performance between DESOM and SOM-CPC is also visible in \cref{fig:audio_som_maps}. SOM-CPC has clustered the SOM nodes belonging to the same speaker, and seems to group male and female speakers (denoted with the node's shape), while these effects are not present in the SOM of the GRU-DESOM model. 

\textbf{Comparison other baselines:}
Minimizing the InfoNCE training objective of CPC with \mbox{$F=2$} was found challenging for this dataset, which resulted in non-competitive performance of the linear classifier and K-means clustering trained on the 2D latent space. Using PCA for dimensionality reduction of the high-dimensional CPC latent space (with $F=512$) performed better, but still inferior to SOM-CPC.

The SOM of the SOM-CPC model (\cref{fig:audio_som_maps}-right) reveals two separate clusters both for speaker 2 (green) and 3 (red). The LibriSpeech corpus contains multiple recordings from each speaker, grouped by (book) chapters from which the speaker was reading. Interestingly, additional analyses revealed that the red and green sub-clusters represented recordings belonging to different chapters: $99.99\%$ of the test-set windows mapped to the upper red sub-cluster belong to the same chapter, while $99.97\%$ of the windows in the lower red sub-cluster belong to another chapter read by this speaker. Similarly, $100.0\%$ of the test set windows in the right green sub-cluster belong to two chapters read by speaker 2, while $98.91\%$ of the windows in the left green sub-cluster belong to another chapter. An auditory inspection revealed that the room acoustics of the recordings belonging to the chapters in different clusters were different, causing changes in the signals which the SOM-CPC model has picked upon. This division between recordings of the same speaker is not visible in the 2D PCA projection of CPC (with $F=128$), see \cref{fig:audio_pca_vs_somcpc} in \cref{app:extended_audio_results}.

\section{Discussion}
\label{sec:discussion}


We formulated a framework that generalizes deep-SOM algorithms, and proposed a new member of this family, SOM-CPC, for interpretable 2D representation learning of high-rate data streams. 

Earlier proposed deep-SOM models used reconstruction objectives. In contrast SOM-CPC leverages contrastive learning and outperformed these models with a wide gap on a variety of metrics. Moreover, it implicitly enforces temporal smoothness, while autoencoder-based models require additional losses and hyperparameter tuning to achieve this. SOM-CPC's task loss (InfoNCE from \citet{Oord2019}) was found to align better with the topological SOM objective than a reconstruction loss, as already hypothesized by \citet{mrabah2020adversarial}. While for some applications CPC could successfully be trained with a 2D latent space directly, optimization was found to be hampered in case of more intricate data spaces. Compared to vanilla CPC, SOM-CPC enables pattern recognition and knowledge discovery. The SOM objective did not hamper CPC optimization. In the synthetic setup it even had a regularizing effect, resulting in lower regression error than vanilla CPC. The use of a Gaussian neighbourhood kernel, as opposed to a plus kernel, was found to improve the topological ordering in the SOM. No decisive conclusions could be made regarding gradient blocking from the SOM loss towards the encoder parameters. Allowing these gradients to flow did not hurt performance, so for coding simplicity, we would advice to not detach the SOM loss.

The model contains several hyperparameters; the latent space dimension $F$, the number of past windows $L$, the number of positive ($P$) and negative ($N$) samples, the number of SOM nodes $k$, the standard deviation of the Gaussian kernel at the end of training, and the loss trade-off parameter $\alpha$ (see \cref{algorithm} in \cref{app:general_details}). Effects of setting different values for $F$ and $\alpha$ were already discussed.

The choice for $L$ and $P$ depends on the expected local stationarity of the signals. CPC assumes that the signal behaves stationary, or at least changes less or in a more predictable manner, within the $L+1+P$ windows around the anchor/current window, compared to the negative windows that are drawn further away in time. The optimal choice for $P$ and $L$ is thus use-case dependent. In contrast, independent of use case, the higher $N$, the more tight the bound on mutual information between the context vector and the future latent space, which is maximized by minimizing the InfoNCE loss \citep{Oord2019}. 

Choosing a large number of SOM nodes $k$ provides more flexibility for clustering purposes, compared to choosing a low number of nodes. The model can learn to make near-by nodes very similar in case the additional flexibility is not needed, however optimization performance might be hampered when choosing an excessive number of nodes. As such we set $k$ to a sensible, but large-enough, value in each experiment. Lastly, setting $\sigma^{(n_{\text{max}})}$ can be more intricate. We observed instable behaviour in the synthetic experiments when the Gaussian kernel became too narrow towards the end of training. Also, its exact value may affect the trained SOM; a low value provides good flexibility in learning specialized codebook vectors at the risk of instability, while a high value induces a higher correlation between these vectors, at the cost of flexibility.

Setting an appropriate stopping criterion for self-supervised models is debatable. In literature, models with the best test set performance are sometimes reported \citep{he2019moco,Fortuin2019}. This is, however, questionable as it may artificially boost reported performance. As such, we created a validation set for model selection in all experiments. Another challenge arises when dealing with aggregated loss functions, since not all losses may smoothly decay and the weighted summation of losses may result in a different optimal epoch than the sub-losses separately. Besides, classification performance does not necessarily perfectly align with SOM performance or information preservation (see \cref{fig:mass_training_curves} in \cref{app:extended_sleep_results}).

Except from determining a suitable stopping criterion, metric computation in unsupervised learning can also be discussed. Section \ref{sec:performance_evaluation} already shortly elucidated upon the way in which we computed classification/regression metrics. Each SOM node was labeled/colored with information from the training set. It can be questioned whether this node labelling should be done on a training or validation set, or directly on the test set for which performance is reported. In earlier times when more conventional (unsupervised) clustering approaches (e.g. K-means) were used, a train/validation/test split was typically not made. As a result, clustering was directly performed on the one and only (test) data set, that was also used to label the clusters/nodes. Moving towards deep learning based approaches where more hyperparameters need to be set and overfitting can become a larger problem, we found it necessary to report on a test set that was not used for labelling the nodes and/or setting hyperparameters. It should thus be taken into account, that direct comparison to results in other deep-clustering works may need a critical eye to see whether similar procedures were used or not.






We believe that SOM-CPC will facilitate knowledge discovery in real-life time series and opens up new research directions for representation learning of time series. Directions include investigation to whether additions like the soft-cluster assignment, cluster hardening loss or a Gaussian latent prior - which have shown to improve the SOM-VAE model \citep{Manduchi2021} - improve SOM-CPC performance as well. Moreover, the CPC objective assumes slowly-changing data characteristics within the time frame in which positive samples are drawn. A multi-modal variational future prediction could possibly improve performance for data that do not meet this assumption. 
\section*{Acknowledgements}
This work was supported by Onera Health and the project `OP-SLEEP'. The project `OP-SLEEP' is made possible by the European Regional Development Fund, in the context of OPZuid. 
\newpage
\bibliographystyle{icml2023}
\bibliography{references, additional_references}


\newpage
\appendix
\onecolumn

\section{Experimental details}
This appendix contains all information for full reproducability of the experiments. Domain-independent details of SOM-CPC and its evaluation are provided in \cref{app:general_details}, while benchmark implementations are discussed in \cref{app:benchmarks}. Domain-specific settings for the experiments on synthetic data, physiological data, and audio experiments are discussed in sections \ref{app:toy_data}, \ref{app:sleep}, and \ref{app:audio}, respectively.

\subsection{General details}
\label{app:general_details}


\Cref{algorithm} provides pseudocode of SOM-CPC, when considering the presence of an AR module. For reproducibility, the full code base can be found at \url{https://github.com/IamHuijben/SOM-CPC.git}.


\begin{algorithm*}[h]
    \centering
    \caption{SOM-CPC}\label{algorithm}
    \begin{algorithmic}
    \small
    \REQUIRE Dataset $\mathcal{X}$, model comprising $f_\theta$, $g_{\psi}$, $\{\mathbf{W}_p\}_{p=1}^P$, $\quant$, latent space dimension $F$, \# past windows $L$, \# positive samples $P$, \#~negative samples $N$, \# SOM nodes $k$, Gaussian neighbourhood function $\mathcal{S}$ with $\sigma^{(n_\text{max})}$, \nmax, loss trade-off parameter $\alpha$.
    \ENSURE A trained SOM: topologically ordered codebook $\Phi$ that represents data $\mathcal{X}$ in 2D. 
    \STATE 
    \STATE - Randomly initialize $\Phi \sim \operatorname{U}[-\sqrt{1/F},\sqrt{1/F}]$
    \STATE - Initialize the Gaussian kernel according to \cref{eq:gaussian_kernel} with: 
    $\sigma^{(0)} = \frac{1}{2} \sqrt{k}$ ~~and~~ $\lambda = -$\nmax$/ \log(\sigma^{(n_{\text{max}})} / \sigma^{(0)})$
    \FOR{$n$ in \nmax}
        \FOR{batch in \# batches}
            \STATE - Sample a sequence of datapoints: $[\vx(t-L),\ldots,\vx(t)] \subset \mathcal{X}$
            \STATE - Define $P$ positive samples: $\{\vx(t+p)\}_{p=1}^P$
            \STATE - Sample $P \times N$ negative samples $\mathcal{X}' \subset \mathcal{X}$, with $|\mathcal{X}'| = P \times N$
            \STATE - Encode: \\
                \begin{itemize}[label={}]
                    \item  \hspace{0.2cm} - data sequence: $\vc(t) = g_{\psi}\Big(f_{\theta}\big([\vx(t-L),\ldots,x(t)]\big)\Big)$
                    \item \hspace{0.2cm} - positive samples: $\{\vz(t+p)\}_{p=1}^P = f_{\theta}\big(\{\vx(t+p)\}_{p=1}^P\big)$
                    \item \hspace{0.2cm} - negative samples: $\mathcal{Z}' = f_{\theta}\big(\mathcal{X}'\big)$
                \end{itemize}
            \STATE - Compute $\mathcal{L}_{\text{topo}}\big(\vc(t)\big)$ and $\mathcal{L}_{\text{infoNCE}}$ according to \cref{eq:topo_loss} and \cref{eq:CPC_loss}
            \STATE - Update trainable parameters $\propto \alpha\mathcal{L}_{\text{topo}}$ and $\mathcal{L}_{\text{infoNCE}}$ 
    \ENDFOR
        \STATE - Update $\sigma^{(n)}$ according to \cref{eq:sigma_over_time}
    \ENDFOR
    \end{algorithmic}
\end{algorithm*}

Several metrics were used to quantify SOM performance, as explained in \cref{sec:performance_evaluation}. The purity implementation was taken from the Github implementation of \citet{Fortuin2019}, while NMI was computed using the sklearn library \citep{scikit-learn}. The topographic error implementation comes from the \textit{SOMperf} python library \citep{forest2020survey}. Finally, the $\ell_{2,\text{smooth}}$ distance is computed using the \textit{norm} function in the \textit{linalg} library of Numpy. 

We here provide the formal definitions of all reported metrics. After SOM training, its nodes are `colored' using a (sub)set of data, which assigns labels $\nu_i$ to each node $\phi_i$ (with $i \in \{1,\ldots,k\}$). When running inference on a data stream $\mX$ and corresponding labels $\mY$, each window $\vx$ in $\mX$ is mapped to a node/codebook vector. If $\vx$ is mapped to node $j$, then the predicted label for $\vx$ equals $\hat{y} = \nu_j$. $\hat{\mY}$ contains the predicted labels for all data points in $\mX$. The definition of all metrics, evaluated over data ($\mX$, $\mY$), are provided below and can straightforwardly be applied for the full dataset ($\mathcal{X}$, $\mathcal{Y}$) by averaging the metrics for each ($\mX$, $\mY$). Purity is defined as:

$$\text{purity} = \frac{1}{|\mX|} \sum_{i=1}^k ~~\sum_{\vx,y \in \mX,\mY} \mathbbm{1}\big(\quant(\vc) = \phi_i ~,~  y=\nu_i \big),$$

with $\mathbbm{1}$ an indicator function that returns one if the condition is met, and zero otherwise. Normalized Mutual Information is defined as:

$$\text{NMI} = \frac{\operatorname{MI}(\mY; \hat{\mY})} {\sqrt{\operatorname{H}(\mY) \times \operatorname{H}(\hat{\mY})}},$$

with $\operatorname{MI}$ being the Mutual Information and $\operatorname{H}$ the Shannon entropy. The regression and classification metrics are:

$$\operatorname{SE}_{\text{target}} = \frac{1}{|\mX|} \sum_{\vx,y \in \mX,\mY} (\hat{y}-y)^2,$$

$$\text{Cohen's Kappa} = \frac{\text{p}_{\text{a}}(\mY, \hat{\mY}) - \text{p}_{\text{c}}(\mY, \hat{\mY}) }{1 - \text{p}_{\text{c}}(\mY, \hat{\mY})},$$

where $\text{p}_{\text{a}}$ and $\text{p}_{\text{c}}$ denote the probability of agreement, and the probability of agreement by chance, respectively. Lastly, metric definitions concerning topological organization are:

$$\ell_{2,\text{smooth}} = \frac{1}{|\mX|} \sum_{\vx \in \mX}\Big|\Big|\mathcal{P}\big[q_{\Phi}\big(\vc(t)\big)\big], \mathcal{P}\big[q_{\Phi}\big(\vc(t+1)\big)\big]\Big|\Big|_2, $$

$$ \operatorname{TE} = \frac{1}{|\mX|} \sum_{\vx \in \mX} \mathbbm{1}\Big(\Big|\Big|\mathcal{P}\big[\quant(\vc)\big], \mathcal{P}\big[\tilde{q}_{\Phi}(\vc) \big]\Big|\Big|_2>1\Big),$$

where $\mathcal{P}$ projects a codebook vector to its corresponding coordinate on the 2D grid, and $\quant(\vc)$ and $\tilde{q}_{\Phi}(\vc)$ denote the winning, respectively, second-best codebook vector for embedding $\vc$. Note that in case the AR module is not used in SOM-CPC all metrics are computed on $\vz$ rather than $\vc$. To enhance readability, time indices $(t)$ were omitted for metrics that do not depend on windows from varying times.

\subsection{Benchmarks and ablations}
\label{app:benchmarks}

To benchmark our implementation of the SOM-VAE model (and the very similar DESOM model that used the same backbone implementation in our code base), we replicated the results on MNIST. MNIST was not used for further experimentation with SOM-CPC in the main body of this paper since this work focuses on high-rate time series. Using $k=16$ SOM nodes, \citet{Fortuin2019} report purity $=0.731\pm0.004$ and NMI $=0.594\pm0.004$ in table 1, and purity $=0.721\pm0.006$ and NMI $=0.587\pm0.003$ in table S1. All numbers are averages and standard errors over 10 runs.

Settings that were not provided in the paper, were taken from the hard-coded settings that we found in the provided code base. Instead of splitting the standard MNIST training set in a training and test split, as done by \citet{Fortuin2019}, we used the available train/test split that comes with the standard MNIST dataloader from Pytorch. From the first ten runs, one run fully collapsed and resulted in extremely poor performance. Considering this run as an outlier, we run an $11^{\text{th}}$ run and here report the average and standard errors of the 10 non-collapsed runs:
purity $=0.705\pm0.002$ and NMI $=0.584\pm0.001$. Our performance does reasonably well match the reported performance by \citet{Fortuin2019}.

To restrict the search space of hyperparameters in the SOM-VAE(-prob) model, which has multiple loss multipliers, we fixed some of the these multipliers for further experiments in this paper. Multiplier $\beta$ scales the SOM loss that sums the quantization error of the four neighbour nodes in the plus kernel. It is, therefore, expected to be at least 4 times larger than the commitment loss, which only reflects the quantization error of the winning node. Choosing $\beta = \alpha/4$ would imply that the weighing of the summed quantization error of the four neighbours is equal to the weighing of this error for the winning node. To give the winning node a slightly higher importance, we set $\beta = \alpha/5 = 0.2\alpha$. The authors of SOM-VAE \citep{Fortuin2019} used, instead, a search strategy to find the optimal setting. Their code base\footnote{\url{https://github.com/ratschlab/SOM-VAE/blob/master/som_vae/somvae_train.py}, line 79.} shows that the SOM loss was multiplied with $0.9$, while $\alpha=1$. Taking into account that their implementation of the commitment loss averaged the quantization error of the four neighbour nodes, while we summed the contribution, their effective setting was thus set to $\beta = \frac{0.9}{4}\alpha = 0.225\alpha$, which is close to what we used in our experiments.

We compared SOM-CPC also against vanilla CPC training followed by a linear classifier or K-means clustering, while freezing the encoder parameters. The supervised linear classifier took in all experiments the form of one fully-connected layer, including biases, that was followed by a log-softmax activation for the physiological and audio cases. It was trained using the mean squared error for the synthetic data set, and cross-entropy loss for physiological and audio experiments. K-means clustering was run from the sklearn library, with the default settings. The number of clusters was chosen to be equal to the number of nodes in the SOM-CPC models against which the performance was compared. Also the disjointly-trained SOM had exactly the same settings as the SOM in the SOM-CPC models with which it was compared. Disjoint training was implemented as follows. First the vanilla CPC model (with $F=128$) was trained, after which the training data was projected into the latent space using the trained encoder. Then the SOM was trained using the loss function in \cref{eq:topo_loss}, while using the fixed latent representations as input.

Several ablations were performed on the SOM-CPC model. We tested the effect of propagating gradients of $\mathcal{L}_{\text{SOM}}$ to the encoder parameters, the difference between using a Gaussian neighbourhoood kernel versus a plus kernel, and the effect of jointly training CPC and the SOM. Moreover, the effect of certain settings that are typically used in the CPC objective are investigated in this appendix. CPC's InfoNCE objective for one window $p$, given in \cref{eq:CPC_loss_per_p}, can as follows be generalized to a more general contrastive learning objective:

\begin{equation}
    \small
    \label{eq:general_loss}
    \mathcal{L}_p = - \underset{\mathcal{X}}{\mathbb{E}} \Big[\log \frac{\exp\Big(\operatorname{sim}(\vz_a, \vz_p)/\tau\Big)}{\sum_{\vz' \in \mathcal{Z}_p'\cup \{\vz_p\}} \exp\Big(\operatorname{sim}(\vz_a, \vz'_p)/\tau\Big)} \Big],
\end{equation}
where $\vz_a$ is the latent space of the current (or \textit{anchor}) window, $\operatorname{sim}(\cdot)$ a similarity metric, $\tau$ a temperature parameter and the other symbols are equivalent to \cref{eq:CPC_loss_per_p}. CPC typically uses $\tau=1$, and the dot product as the similarity metric. However, other related contrastive learning objectives, e.g. in SimCLR \citep{Chen2020}, use a temperature value that is often set to 0.07 \citep{Chen2020, Woo2022CoST:Forecasting}, and a cosine similarity instead of the (unnormalized) dot product. As such, in all experiments we added ablations where we set $\tau$ at 1 or 0.07, and use either the dot-product or the cosine similarity as the similarity metric. Results are discussed in \cref{app:toy_extended_res,app:extended_sleep_results,app:extended_audio_results}.

\subsection{Synthetic experiments}
\label{app:toy_data}

\subsubsection{Training details}
\label{app:toy_architecture}

Table \ref{tab:toy_model_details} summarizes the encoder and decoder architectures used in this experiment. The \textit{output size} column in the table uses channels-first notation. The SOM-VAE and DESOM models were found to benefit from a convolutional part of the encoder that did not fully reduce the temporal dimension to size 1. As such, the last convolutional layer of the SOM-CPC encoder was changed for a fully connected layer preceded by a flattening operation for the autoencoder-based models. The SOM-CPC and CPC model are run without an AR module, to make the fairest comparison to the SOM-VAE(prob) and DESOM models, which also do not incorporate such a component.


\begin{table}
\centering
\tiny
\caption{Model details for the synthetic data experiments in \cref{sec:synthetic_data}.}
\label{tab:toy_model_details}
\begin{tabular}{llcccccc}
\textbf{Layer type} & \textbf{Output size} & \textbf{Channels} & \textbf{Activation} & \textbf{Kernel size} & \textbf{Strides} & \textbf{Dilation} &\textbf{Padding} \\
\hline\hline
\rowcolor{Gray}
\multicolumn{8}{l}{\textbf{\hspace{-0.2cm}Encoder for SOM-CPC and CPC}}\\\hline
Conv1D  & bs $\times 16 \times 128$ & 16 & Leaky ReLU (0.01)  & $9$  & $1$ & $1$ & same \\[0.1em]
MaxPool1D & bs $\times 16 \times 32$ & - & - & $4$ & $4$ & - & -\\[0.1em]
Dropout ($0.1$) & bs $\times 16 \times 32$ & - & - & - & - & - & -\\[0.1em]
Conv1D  & bs $\times 32 \times 32$ & 32 & Leaky ReLU (0.01)  & $7$  & $1$ & $1$ & same \\[0.1em]
MaxPool1D &  bs $\times 32 \times 8$ & - & - & $4$ & $4$ & - & -\\[0.1em]
Dropout ($0.1$) & bs $\times 32 \times 8$ & - & - & - & - & - & -\\[0.1em]
Conv1D  &  bs $\times 64 \times 8$ & 64 & Leaky ReLU (0.01)  & $3$  & $1$ & $1$ & same \\[0.1em]
MaxPool1D &  bs $\times 64 \times 2$ & - & - & $4$ & $4$ & - & -\\[0.1em]
Dropout ($0.1$) & bs $\times 64 \times 2$ & - & - & - & - & - & -\\[0.1em]
Conv1D  & bs $\times 128 \times 2$ & 128 & Leaky ReLU (0.01)  & $3$  & $1$ & $1$ & same \\[0.1em]
MaxPool1D & bs $\times 128 \times 1$ & - & - & $2$ & $2$ & - & -\\[0.1em]\hline
\rowcolor{Gray}
\multicolumn{8}{l}{\textbf{\hspace{-0.2cm}Encoder for SOM-VAE(-prob) and DESOM}}\\\hline
Conv1D  & bs $\times 16 \times 128$ & 16 & Leaky ReLU (0.01)  & $9$  & $1$ & $1$ & same \\[0.1em]
MaxPool1D & bs $\times 16 \times 32$ & - & - & $4$ & $4$ & - & -\\[0.1em]
Dropout ($0.1$) & bs $\times 16 \times 32$ & - & - & - & - & - & -\\[0.1em]
Conv1D  & bs $\times 32 \times 32$ & 32 & Leaky ReLU (0.01)  & $7$  & $1$ & $1$ & same \\[0.1em]
MaxPool1D &  bs $\times 32 \times 8$ & - & - & $4$ & $4$ & - & -\\[0.1em]
Dropout ($0.1$) & bs $\times 32 \times 8$ & - & - & - & - & - & -\\[0.1em]
Conv1D  &  bs $\times 64 \times 8$ & 64 & Leaky ReLU (0.01)  & $3$  & $1$ & $1$ & same \\[0.1em]
Flatten & bs $\times 512$ & - & - & - & - & - & - \\[0.1em]
Fully Connected & bs $\times 128$ & 128 & Leaky ReLU (0.01) & - & - & - & -\\[0.1em]
\hline
\rowcolor{Gray}
\multicolumn{8}{l}{\textbf{\hspace{-0.2cm}Decoder for SOM-VAE(-prob) and DESOM}}\\\hline
Fully Connected & bs $\times 512$ & 512 & Leaky ReLU (0.01) & - & - & - & -\\[0.1em]
Unflatten & bs $\times 64 \times 8$ & - & - & - & - & - & - \\[0.1em]
Conv1D  & bs $\times 32 \times 8$ & 32 & Leaky ReLU (0.01)  & $3$  & $1$ & $1$ & same \\[0.1em]
ConvTranspose1D  & bs $\times 32 \times 32$ & 32 & None & $4$  & $4$ & $1$ & $0$ \\[0.1em]
Conv1D  & bs $\times 16 \times 32$ & 16 & Leaky ReLU (0.01)  & $7$  & $1$ & $1$ & same \\[0.1em]
ConvTranspose1D  & bs $\times 16 \times 128$ & 16 & None & $4$  & $4$ & $1$ & $0$ \\[0.1em]
Conv1D  & bs $\times 1 \times 128$ & 1 & Tanh & $9$  & $1$ & $1$ & same \\\hline
\end{tabular}
\end{table}

For SOM-CPC, $P=3$ future predictions (i.e. positive samples) were used, and $N = 3$ negative samples were drawn for each positive sample. The latter were drawn randomly from the entire training set. The standard deviation of the Gaussian neighbourhood kernel was exponentially decayed until $\sigma^{(n_{\text{max}})} = 2$. Choosing a lower value at the end of training induced instable optimization behaviour.

All models (including the benchmarks) were trained using the Adam optimizer \citep{kingma2014adam}, with a learning rate of 0.001 and a batch size of 128. Each model was trained for maximally 1000 epochs. The best model was selected based on the lowest task loss on the validation set, being $\mathcal{L}_{\text{recon}}$ for SOM-VAE and DESOM and $\mathcal{L}_{\text{InfoNCE}}$ for SOM-CPC and CPC. We did not use the full training objective $\mathcal{L}_{\text{deep-SOM}}$ as model selection criterion, as both the commitment and SOM loss showed to be low initially (possibly due to low values of the random initialization of the model), while both increased and reached a steady-state later in training. The linear classifier and the disjointly-trained SOM  on the CPC embeddings were trained until convergence, for maximally 1000 epochs.

\subsubsection{Extended results}
\label{app:toy_extended_res}

Table \ref{tab:toy_results_extended} extends \cref{tab:toy_results} with additional sweeps of hyperparameters $\alpha$, $\gamma$, and $\zeta$, and ablations with different settings of the temperature value $\tau$ and the used similarity metric. Compared to SOM-VAE, SOM-VAE-prob is expected to show more smooth trajectories over time, captured in the $\ell_{2,\text{smooth}}$ distance metric,  thanks to the additional transition and smoothness loss (multiplied by $\gamma$ and $\zeta$, respectively). It can be seen that tuning these hyperparameters is a complex process, and a sweep did not result in one SOM-VAE-prob run that performed better than the best SOM-VAE model. In contrary, the addition of the extra losses possibly interfered with the optimization process, and only very delicate settings of $\gamma$ and $\zeta$ might improve model performance eventually. A sweep over the topological loss multiplier $\alpha$, revealed a low sensitivity of SOM-CPC to this value. It can be seen that changing the temperature value and/or the similarity metric did not significantly alter the performance consistently on all reported metrics.

\begin{figure}[t]
    \centering
    \includegraphics[width=1\linewidth,trim={0cm 13.5cm 10.5cm 2.3cm},clip]{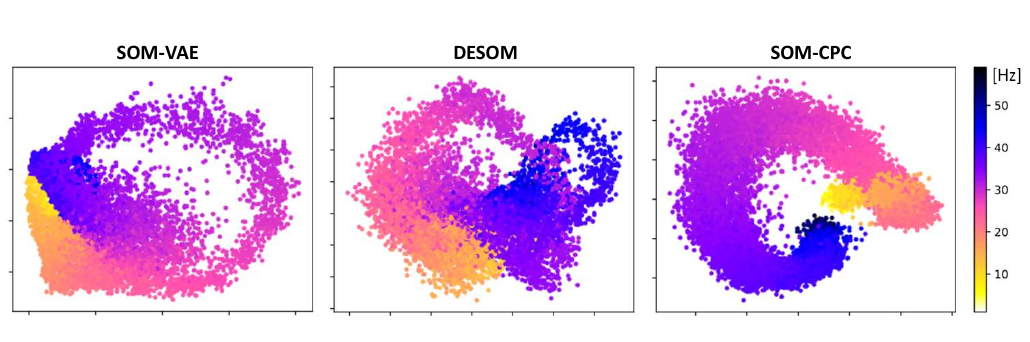}
    \\[-0.5em]
    \caption{PCA projections of the continuous latent spaces of the full test set for the SOM-VAE, DESOM, and SOM-CPC models of which the SOMs were visualized in \cref{fig:som_maps_toy}. Organization of the signal frequencies is most structured in the SOM-CPC model.}    
    \label{fig:PCA_toy}
\end{figure}

Figure \ref{fig:PCA_toy} shows PCA projections of the latent space of the SOM-VAE, DESOM, and SOM-CPC models that are indicated with a * in \cref{tab:toy_results,tab:toy_results_extended}, and for which the SOMs were visualized in \cref{fig:som_maps_toy}. Organization of the signal frequencies is much better for the SOM-CPC model, providing an explanation for the better (i.e. lower) $\operatorname{SE}_{\text{target}}$ of this model, as compared to SOM-VAE and DESOM.

\begin{table}[t]
    \caption{This is the extended version of \cref{tab:toy_results} on the synthetic data results, including a sweep of hyperparameters $\alpha$, $\gamma$ and $\zeta$. Bold values indicate the best performance per column (excluding the upper bound of the vanilla CPC model, which does not result in a 2D representation). The models indicated with a $*$ were used to depict trained SOMs in \cref{fig:som_maps_toy} and PCA projections in \cref{fig:PCA_toy}.}
    \label{tab:toy_results_extended}
\centering
   \tiny
   \begin{tabular}{p{0.03cm}lccc|lll}
    	\hline
    	\rowcolor{Gray}
    	& \textbf{Model} & $\boldsymbol{\alpha}$ & $\mathcal{S}$ & $\mathcal{L}_{\text{SOM}} \operatorname{sg}[\cdot]$ &  $\operatorname{SE}_{\text{target}}$ & \textbf{$\ell_{2,\text{smooth}}$}  & \textbf{TE} \\\hline\hline
    	& CPC + linear classifier & - & - & - & ~~2.62$\pm$~~2.37 & - & -
    	\\
    	& CPC + K-means & - & - & - & ~~1.09$\pm$~~~~.62  & - & - 
    	\\\hline
    	& CPC ($F=2$)~+ linear classifier & - & - & - & 25.01$\pm$42.94
        & - & - 
        \\
        & CPC ($F=2$)~+ K-means & - & - & - & ~~~~.76$\pm$~~1.31
        & - & - 
    	\\
    	& CPC + PCA + linear classifier & - & - & - & 42.81$\pm$58.12  & - & - 
    	\\
    	& CPC + PCA + K-means & - & - & - & ~~4.42$\pm$~~9.01  & - & - 
    	\\\hline
        & SOM-VAE  & 1e-3 & Plus & \cmark & 11.59$\pm$13.69	&	2.60$\pm$.46	&	.38$\pm$.04
        \\
        & & 1e-2 & Plus & \cmark & 14.57$\pm$44.10	&	2.98$\pm$.84	&	.38$\pm$.05
        \\
        * & & .1   & Plus &  \cmark  & ~~8.02$\pm$~~4.58	&	2.41$\pm$.68	&	.28$\pm$.07
        \\
        &  & 1 & Plus & \cmark & 13.43$\pm$~~3.66	&	2.75$\pm$.45	&	.33$\pm$.05
        \\\hdashline
        & SOM-VAE & 1e-5 & Gaussian & \cmark  & 11.12$\pm$17.05	&	1.93$\pm$.29	&	.07$\pm$.03
        \\
        & & 1e-4 & Gaussian & \cmark  & ~~1.52$\pm$26.61	&	1.95$\pm$.36	&	.08$\pm$.03
        \\
        & & 1e-3 & Gaussian & \cmark  & 11.60$\pm$25.43	&	1.92$\pm$.34	&	.06$\pm$.02
        \\ 
        &  & 1e-2 & Gaussian & \cmark  & 18.13$\pm$48.66	&	2.03$\pm$.42	&	.09$\pm$.03
        \\\hline
        & SOM-VAE-prob \hspace{0.55cm}($\gamma = \mbox{5e-5}$, $\zeta=\mbox{1e-3}$) & 1e-3 & Plus & \cmark  & 21.26$\pm$55.00	&	3.78$\pm$.52	&	.93$\pm$.04

        \\ 
        &  \hspace{2.0cm}($\gamma = \mbox{4e-5}$, $\zeta=\mbox{1e-3}$)  & 1e-3 & Plus & \cmark & 21.30$\pm$37.37	&	4.23$\pm$.57	&	.88$\pm$.05

        \\ 
        & \hspace{2.0cm}($\gamma = \mbox{3.3e-5}$, $ \zeta=\mbox{1e-3}$) & 1e-3 & Plus & \cmark & 22.52$\pm$48.97	&	3.81$\pm$.51	&	.98$\pm$.02

        \\ 
        & \hspace{2.0cm}($\gamma = \mbox{5e-4}$, $\zeta=\mbox{1e-2}$) & 1e-2  & Plus & \cmark  &  14.65$\pm$19.58	&	3.70$\pm$.51	&	.86$\pm$.08

        \\ 
        & \hspace{2.0cm}($\gamma = \mbox{4e-4}$, $\zeta=\mbox{1e-2}$) & 1e-2  & Plus & \cmark  & 27.25$\pm$72.82	&	3.54$\pm$.67	&	.94$\pm$.02

        \\ 
        & \hspace{2.0cm}($\gamma = \mbox{3.3e-4}$, $\zeta=\mbox{1e-2}$)& 1e-2 & Plus & \cmark  & 21.62$\pm$38.78	&	3.71$\pm$.76	&	.97$\pm$.01

        \\ 
        & \hspace{2.0cm}($\gamma = \mbox{5e-5}$, $ \zeta=\mbox{1e-2}$) & .1  & Plus & \cmark & 26.82$\pm$68.84	&	3.23$\pm$.80	&	.68$\pm$.07

        \\ 
        & \hspace{2.0cm}($\gamma = \mbox{4e-5}$, $\zeta=\mbox{1e-2}$) & .1 & Plus & \cmark & 22.34$\pm$64.92	&	3.11$\pm$.73	&	.74$\pm$.07

        \\ 
        & \hspace{2.0cm}($\gamma = \mbox{3.3e-5}$, $\zeta=\mbox{1e-2}$) &.1 & Plus & \cmark & ~~2.10$\pm$48.80	&	3.15$\pm$.73	&	.63$\pm$.05

        \\ 
        & \hspace{2.0cm}($\gamma = \mbox{5e-4}$, $\zeta=\mbox{.1}$)  &.1 & Plus & \cmark & ~~4.85$\pm$75.96	&	3.06$\pm$.55	&	.84$\pm$.05

        \\
        & \hspace{2.0cm}($\gamma = \mbox{4e-4}$, $\zeta=\mbox{.1}$)  & .1 & Plus & \cmark & 29.96$\pm$46.96	&	2.81$\pm$.34	&	.82$\pm$.13

        \\
        & \hspace{2.0cm}($\gamma = \mbox{3.3e-4}$, $ \zeta=\mbox{.1}$) & .1  & Plus & \cmark & ~~3.96$\pm$56.94	&	3.95$\pm$.83	&	.85$\pm$.11
        \\\hline
        & DESOM & 1e-5 & Gaussian & \xmark & 14.00$\pm$~~3.10	&	1.99$\pm$.36	&	.13$\pm$.07
        \\
        &  & 1e-4 & Gaussian & \xmark & 19.09$\pm$44.04	&	1.95$\pm$.28	&	.11$\pm$.05
        \\
        &  & 1e-3 & Gaussian & \xmark & 12.58$\pm$27.10	&	1.92$\pm$.31	&	.08$\pm$.03
        \\
        &  & 1e-2 & Gaussian & \xmark & 13.66$\pm$44.71	&	1.89$\pm$.33	&	.07$\pm$.03
        \\
         * &  & .1 & Gaussian & \xmark & 10.77$\pm$10.85	&	2.20$\pm$.44	&	.06$\pm$.02
         \\
         &  & 1 & Gaussian & \xmark & 22.86$\pm$55.71	&	2.26$\pm$.43	&	.09$\pm$.03
        \\\hline
         & SOM-CPC (ours) & 1e-5 & Gaussian & \xmark & ~~~~.95$\pm$~~2.40	&	1.24$\pm$.31	&	.05$\pm$.02
         \\
         * & & 1e-4 & Gaussian & \xmark &~~~~.72$\pm$~~1.08	&	1.37$\pm$.37	& .02$\pm$.01
         \\
         & & 1e-3 & Gaussian & \xmark & ~~~~.81$\pm$~~~~.75	&	1.06$\pm$.28	&	.03$\pm$.01
          \\
          & & 1e-2 & Gaussian & \xmark & ~~~~.62$\pm$~~~~.71	&	1.08$\pm$.28	&	.06$\pm$.04
         \\
          & & .1 & Gaussian & \xmark &~~1.90$\pm$~~4.07	&	1.18$\pm$.30	&	.04$\pm$.02
         \\\hdashline
         \hspace{-0.5cm}\parbox[t]{2mm}{\multirow{11}{*}{\rotatebox[origin=c]{90}{\underline{~~~~~~~~~~~~~~Ablations~~~~~~~~~~~~~~}} }} &  & 1e-5 & Gaussian & \cmark & ~~~~.64$\pm$~~~~.71	&	1.04$\pm$.26	&	.05$\pm$.02

         \\
         & & 1e-4 & Gaussian & \cmark & ~~~~.68$\pm$~~~~.67	&	1.19$\pm$.43	&	.06$\pm$.03

         \\
         & & 1e-3 & Gaussian & \cmark & ~~~~.75$\pm$~~1.02	&	1.12$\pm$.32	&	.06$\pm$.03

         \\
        & & 1e-2 & Gaussian & \cmark & \textbf{~~~~.47$\pm$~~~~.48}	&	\textbf{~~.99$\pm$.24}	&	.07$\pm$.04
        \\
        & & .1 & Gaussian & \cmark & ~~~~.89$\pm$~~1.50	&	1.16$\pm$.32	&	.07$\pm$.03
         \\
   	    & & 1e-4 & Plus & \xmark & ~~1.71$\pm$~~1.15	&	2.46$\pm$.51	&	.30$\pm$.06
   	    \\
    	& & 1e-2 & Plus & \cmark  & ~~1.16$\pm$~~~~.61	&	1.85$\pm$.26	&	.12$\pm$.04
        \\
        & SOM-CPC ($\tau=0.07$, $\operatorname{sim} =$ cosine sim.)& 1e-4 & Gaussian & \xmark & ~~1.47$\pm$~~2.60 &	1.15$\pm$.34 & \textbf{.01$\pm$.02} \\
        & SOM-CPC ($\tau=1$, $\operatorname{sim} =$ cosine sim.) & 1e-4 & Gaussian & \xmark & ~~2.26$\pm$~~4.91 &	~~.96$\pm$.13	&.06$\pm$.02 \\
        & SOM-CPC ($\tau=0.07$, $\operatorname{sim} =$ dot prod.) & 1e-4 & Gaussian & \xmark & ~~1.22$\pm$~~4.08 &	1.15$\pm$.37 & .07$\pm$.05 \\
        & CPC + SOM (disjoint) & - & Gaussian & - & ~~~~.84$\pm$~~1.14 & 1.47$\pm$.50 & .03$\pm$.01 
    	\\\hline
    \end{tabular}
\end{table}


\clearpage
\subsection{Physiological data experiments}
\label{app:sleep}

\subsubsection{Data processing}
\label{app:sleep_data_preprocessing}

For the experiments on physiological data, we used subset 3 of the publicly available Montreal Archive of Sleep Studies (MASS) database \citep{OReilly2014}, consisting of 62 whole-night polysomnography recordings. Each recording contains, among others, electroencephalography (EEG), chin electromyography (EMG), and electrooculography (EOG) data. We refer the reader to \cite{OReilly2014} for more details regarding this dataset. We selected the channels that are typically used in clinical practice, comprising three EEG channels (F4, C4, O2), the two EOG channels, and one chin EMG derivation, and downsampled the data to 128 Hz. Sleep stage labels that follow the guidelines of the American Academy of Sleep Medicine (AASM) \cite{berry2012aasm} (being wakefulness (Wake), rapid-eye movement (REM) sleep, or non-REM1 till non-REM3 (N1, N2, N3)) were available for every non-overlapping 30-second window, the common label resolution in clinical practice. 

Some processing had already been done by the distributors of the MASS dataset \citep{OReilly2014}. The 60 Hz powerline interference was, however, not fully suppressed, and we wanted to down sample each signal to 128 Hz to reduce computational complexity. As such, before downsampling, all derivations were additionally filtered with a zero-phase (i.e. two-directional) $5^\text{th}$ order Butterworth band-pass filter ($0.3-59$ Hz), followed by another zero-phase $5^\text{th}$ order Butterworth notch filter ($59-61$ Hz). Channels were normalized within-patient and per channel, yielding mean subtraction, followed by normalization such that amplitudes of 95\% of the samples were mapped between -1 and +1. The 62 recordings (numbered $1-64$, with number 43 and 49 missing) were split into a training set including patients $1-48$ ($n=47$), a validation set including patients $50-57$ ($n=8$), and hold-out test set that included patients $58-64$ ($n=7$).

\subsubsection{Training details}
\label{app:sleep_architectures}

Dimensionality reduction of polysomnography data was done by encoding all selected channels in each non-overlapping 30-second window using standard convolutional encoder. Table \ref{tab:sleep_model_details} summarizes the used encoder and decoder (for SOM-VAE and DESOM) architectures. The latent space for decoding in the SOM-VAE and DESOM benchmark models was not fully reduced to a 1D vector to enhance training. Nevertheless, the last adaptive average pooling layer that was used in the encoder of SOM-CPC, was applied in the bottleneck of SOM-VAE and DESOM before SOM quantization took place. As a result the codebook vectors in all models were of size $F=128$. The decoder architecture (see \cref{tab:sleep_model_details}) was used both for the continuous and discrete decoding in the SOM-VAE model (without weight tying). No AR-component was used in the SOM-CPC model to make a fair comparison to the SOM-VAE and DESOM model that also did not include such a component.

For the SOM-CPC model, $P=3$ future predictions (i.e. positive samples) were used, and $N = 3$ negative samples were drawn for each positive sample. The latter were drawn from the same subject as the positive sample. The $\sigma$ of the Gaussian neighbourhood kernel was exponentially annealed to $\sigma^{(n_\text{max})} = 0.5$ during training.

All models were trained with the Adam optimizer \citep{kingma2014adam}, with a learning rate of $1\text{e-}4$ and a batch size of 128. Each model was trained for maximally 500 epochs, and the best model was selected based on the lowest $\mathcal{L}_{\text{recon}}$ (for SOM-VAE and DESOM) or $\mathcal{L}_{\text{InfoNCE}}$ (for SOM-CPC and CPC) on the validation set. 


\begin{table}
\centering
\tiny
\caption{Model details for the experiments on physiological data in \cref{sec:modelling_sleep}.}
\label{tab:sleep_model_details}
\begin{tabular}{llcccccc}
\textbf{Layer type} & \textbf{Output size} & \textbf{Channels} & \textbf{Activation} & \textbf{Kernel size} & \textbf{Strides} & \textbf{Dilation} &\textbf{Padding} \\
\hline\hline
\rowcolor{Gray}
\multicolumn{8}{l}{\textbf{\hspace{-0.2cm}Encoder for SOM-CPC and CPC}}\\\hline
Conv1D  & bs $\times 16 \times 3826$ & 16 & Leaky ReLU (0.01)  & $15$  & $1$ & $1$ & 0 \\[0.1em]
MaxPool1D & bs $\times 16 \times 765$ & - & - & $5$ & $5$ & - & -\\[0.1em]
Dropout ($0.1$) & bs $\times 16 \times 765$ & - & - & - & - & - & -\\[0.1em]
Conv1D  & bs $\times 32 \times 757$ & 32 & Leaky ReLU (0.01)  & $9$  & $1$ & $1$ & 0 \\[0.1em]
MaxPool1D &  bs $\times 32 \times 151$ & - & - & $5$ & $5$ & - & -\\[0.1em]
Dropout ($0.1$) & bs $\times 32 \times 151$ & - & - & - & - & - & -\\[0.1em]
Conv1D  &  bs $\times 64 \times 147$ & 64 & Leaky ReLU (0.01)  & $5$  & $1$ & $1$ & 0 \\[0.1em]
MaxPool1D &  bs $\times 64 \times 29$ & - & - & $5$ & $5$ & - & -\\[0.1em]
Dropout ($0.1$) & bs $\times 64 \times 29$ & - & - & - & - & - & -\\[0.1em]
Conv1D  & bs $\times 128 \times 27$ & 128 & Leaky ReLU (0.01)  & $3$  & $1$ & $1$ & 0 \\[0.1em]
AdaptiveAvgPool1D & bs $\times 128 \times 1$ & - & - & - & - & - & - \\[0.1em]
\hline
\rowcolor{Gray}
\multicolumn{8}{l}{\textbf{\hspace{-0.2cm}Encoder for SOM-VAE and DESOM}}\\\hline
Conv1D  & bs $\times 16 \times 3861$ & 16 & Leaky ReLU (0.01)  & $15$  & $1$ & $1$ & ($18, 17$) \\[0.1em]
MaxPool1D & bs $\times 16 \times 772$ & - & - & $5$ & $5$ & - & -\\[0.1em]
Dropout ($0.1$) & bs $\times 16 \times 772$ & - & - & - & - & - & -\\[0.1em]
Conv1D  & bs $\times 32 \times 764$ & 32 & Leaky ReLU (0.01)  & $9$  & $1$ & $1$ & 0 \\[0.1em]
MaxPool1D &  bs $\times 32 \times 152$ & - & - & $5$ & $5$ & - & -\\[0.1em]
Dropout ($0.1$) & bs $\times 32 \times 152$ & - & - & - & - & - & -\\[0.1em]
Conv1D  &  bs $\times 64 \times 148$ & 64 & Leaky ReLU (0.01)  & $5$  & $1$ & $1$ & 0 \\[0.1em]
MaxPool1D &  bs $\times 64 \times 29$ & - & - & $5$ & $5$ & - & -\\[0.1em]
Dropout ($0.1$) & bs $\times 64 \times 29$ & - & - & - & - & - & -\\[0.1em]
Conv1D  & bs $\times 128 \times 27$ & 128 & Leaky ReLU (0.01)  & $3$  & $1$ & $1$ & 0 \\[0.1em]
\hline
\rowcolor{Gray}
\multicolumn{8}{l}{\textbf{\hspace{-0.2cm}Decoder for SOM-VAE and DESOM}}\\\hline
Conv1D  & bs $\times 64 \times 27$ & 64 & Leaky ReLU (0.01)  & $3$  & $1$ & $1$ & 1 \\[0.1em]
ConvTranspose1D  & bs $\times 64 \times 135$ & 64 & None  & $5$  & $5$ & $1$ & 0 \\[0.1em]
Conv1D  & bs $\times 32 \times 135$ & 32 & Leaky ReLU (0.01)  & $5$  & $1$ & $1$ & 2 \\[0.1em]
ConvTranspose1D  & bs $\times 32 \times 675$ & 32 & None  & $5$  & $5$ & $1$ & 0 \\[0.1em]
Conv1D  & bs $\times 16 \times 675$ & 16 & Leaky ReLU (0.01)  & $9$  & $1$ & $1$ & 4 \\[0.1em]
ConvTranspose1D  & bs $\times 16 \times 3375$ & 16 & None  & $5$  & $5$ & $1$ & 0 \\[0.1em]
Conv1D  & bs $\times 6 \times 3375$ & 6 & None  & $15$  & $1$ & $1$ & 7 \\[0.1em]
Crop & bs $\times 6 \times 3340$ & - & - & - & - & - & - \\[0.1em]\hline
\end{tabular}
\vspace{-0.5cm}
\end{table}

\subsubsection{Extended Results \label{app:extended_sleep_results}}
Table \ref{tab:sleep_results} shows the quantitative results on physiological data, comparing SOM-CPC against deep-SOM models (SOM-VAE and DESOM) and disjoint training of CPC, followed by either a supervised linear classifier, K-means or a SOM. Discussion of the main results in this table can be found in \cref{sec:modelling_sleep}. The ablation experiments in which the value of $\tau$ and/or the similarity metric was altered show that classification and clustering performance slightly dropped when using the cosine similarity with a temperature value of 1, while the topographic organization slightly improved (i.e. lower TE). These effects vanished when using a temperature of $\tau=0.07$. Both runs showed worse temporal smoothness (i.e. higher $\ell_{2,\text{smooth}}$). As expected, only changing the temperature value to 0.07 did almost not affect results, suggesting that the linear projector heads were able to adjust for this scaling factor.

We also tested the performance when using the SimCLR \citep{Chen2020} objective for the task loss, instead of the CPC objective. SimCLR is also a contrastive learning framework, but instead of drawing positive samples from the future latent space, these samples are created by applying augmentations on the anchor window. Inspired by \cite{um2017data} we used the following augmentations: independent and identically distributed Gaussian noise $\mathcal{N}(0,0.05)$ was added (called jitter in their implementation), each channel was scaled with a value drawn from $\mathcal{N}(0,0.1)$, windows were split in 4 sub-windows of minimal 2 seconds and randomly permuted, and lastly time series were both time warped and magnitude warped. The latter two augmentations make use of smooth curves that smoothly vary the positions of time stamps or magnitude values, respectively.

Besides the difference on how to create positive samples, the originally proposed SimCLR model has some other slight differences with respect to the CPC model:

\begin{itemize}
\item The SimCLR loss uses the cosine similarity, while CPC uses the (unnormalized) dot product as the similarity metric (see \cref{eq:general_loss}).
\item SimCLR uses an additional temperature $\tau$ in its loss function  (see \cref{eq:general_loss}), for which the value is often set to 0.07 \citep{Chen2020, Woo2022CoST:Forecasting}. CPC does not incorporate such a temperature, which effectively means that it uses a value of 1.
\item SimCLR uses a non-linear MLP projection head, while CPC uses linear projection heads.
\item	SimCLR uses negative samples from within the batch, while this is not specified in the CPC paper. This specified design choice makes SimCLR typically very sensitive to the batch size.
\item SimCLR was not proposed to include an auto-regressive component, and can not straightforwardly be extended to do so, while CPC can be implemented with or without such a module. 
\end{itemize}

For the most fair comparison, the procedure for drawing negative samples in SimCLR is done equivalently as for SOM-CPC, i.e. within the recording, instead of within the batch. However, in the SOM-SimCLR model (i.e. the joint training of SOM with SimCLR), each drawn negative sample is added to the set of negative samples both in its raw form, and with a random augmentation, which effectively doubles the number of negative samples. Table \ref{tab:sleep_results} reports the performance of SOM-SimCLR (i.e. jointly optimizing SimCLR for feature extraction and a SOM), both for $\tau = 0.07$ and $\tau = 1$, while using the cosine similarity. All settings regarding training procedure and the SOM were set equivalently as in the SOM-CPC training. Table \ref{tab:sleep_results} shows that SOM-SimCLR results for $\tau=0.07$ are better than those with $\tau=1$, which is in line with findings from \cite{Chen2020, Woo2022CoST:Forecasting}.  However, even with $\tau=0.07$, performance of SOM-SimCLR is lower on all metrics compared to the SOM-CPC model with the same value for $\alpha$. The higher $\ell_{2,\text{smooth}}$ metric of SOM-SimCLR indicates on average larger jumps over the SOM map through time, which might be caused by the fact that the SimCLR task objective does not incorporate temporal information, while InfoNCE does exploit this. Training time of SOM-SimCLR was, moverover, considerably longer than SOM-CPC with the same settings due to the additional augmentations that need to be computed for every data window and its negative samples.

\begin{figure}[]
    \centering
    \includegraphics[width=1\linewidth,trim={0cm 0cm 0cm 0cm},clip]{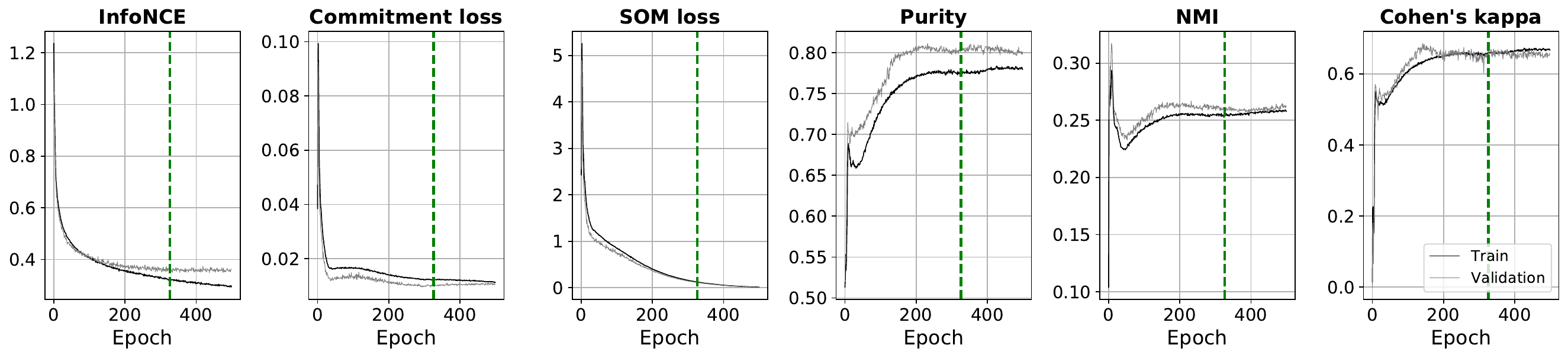}
    \caption{Training curves of SOM-CPC (the model indicated with a * in \cref{tab:sleep_results}) for both the training and validation set. The green dashed line indicates the epoch of the used model, i.e. the one with the lowest validation InfoNCE loss. It can be seen that the epoch with the best clustering and classification performance does not necessarily align with the epoch that has the lowest loss commitment and/or SOM loss.}
    \label{fig:mass_training_curves}
\end{figure}

\Cref{fig:mass_training_curves} shows training curves of the training and validation set for the SOM-CPC model that is indicated with a * in \cref{tab:sleep_results}. The green line indicates the epoch with the lowest InfoNCE validation loss. These graphs show that the performance of InfoNCE, the commitment and SOM loss, and classification metrics do not necessarily align, making it dependent on your final goal with the SOM-CPC model what is the most appropriate stopping-criterion.

\begin{table}[]
\caption{Test set performance of various models trained on physiological recordings. SOMs of models with a * are visualized in \cref{fig:sleep_som_maps}. Bold values indicate the best performance per column (excluding the upper bound of the vanilla CPC model, which does not result in a 2D representation).}
\label{tab:sleep_results}
\centering
   \tiny
   \begin{tabular}{p{0.1cm}lccc|lllll}
    	\hline
    	\rowcolor{Gray}
    	& \textbf{Model} & $\boldsymbol{\alpha}$ & $\mathcal{S}$ & $\mathcal{L}_{\text{SOM}} \operatorname{sg}[\cdot]$ &  \textbf{Purity} & \textbf{NMI} & \textbf{Cohen's kappa} & \textbf{$\ell_{2,\text{smooth}}$} & \textbf{TE}
    	\\\hline\hline
    	& CPC + linear classifier & - & - & - & - & - &  .68$\pm$.10 & - &- \\
    	& CPC + K-means & - & - & - & .79	&	.29	&	.61$\pm$.11 & - & -
    	\\\hline
    	& CPC ($F=2$)~+ linear classifier & - & - & - & -	&	-	&	.52$\pm$.10 & - & - 
    	\\
    	& CPC ($F=2$)~+ K-means & - & - & - & .74	&	.24	&	.55$\pm$.09 & - & -
    	 \\
    	& CPC + PCA + linear classifier & - & - & - & -	&	-	&.54$\pm$.09& - & - 
    	\\
    	* & CPC + PCA + K-means & - & - & - & .77	&	.26	&	.57$\pm$.08 & - & - 
 \\\hline
        & SOM-VAE & 1e-3   & Plus & \cmark  & .71	&	.23	&	.51$\pm$.04	&	2.36$\pm$.26	&	.24$\pm$.03
        \\
        &  & 1e-2   & Plus &  \cmark  & .71	&	.23	&	.51$\pm$.04	&	2.67$\pm$.17	&	.30$\pm$.04
        \\
        &  & .1   & Plus &  \cmark  & .72	&	.23	&	.52$\pm$.03	&	2.60$\pm$.34	&	.28$\pm$.04
        \\
        &  & 1   & Plus &  \cmark  & .71	&	.23	&	.53$\pm$.03	&	3.08$\pm$.32	&	.31$\pm$.05
        \\\hline 
        & DESOM & 1e-6 & Gaussian & \xmark & .70	&	.27	&	.53$\pm$.05	&	2.14$\pm$.32	&	.10$\pm$.02
        \\
        &  & 1e-5 & Gaussian & \xmark & .70	&	.23	&	.50$\pm$.04	&	2.10$\pm$.26	&	.11$\pm$.03
        \\
        &  & 1e-4 & Gaussian & \xmark & .71	&	.22	&	.51$\pm$.04	&	2.35$\pm$.24	&	.17$\pm$.04
        \\
        &  & 1e-3 & Gaussian & \xmark & .71	&	.22	&	.51$\pm$.05	&	2.40$\pm$.16	&	.22$\pm$.01
        \\
        &  & 1e-2 & Gaussian & \xmark & .71	&	.22	&	.50$\pm$.04	&	2.30$\pm$.26	&	.23$\pm$.02
\\\hline
        & SOM-SimCLR ($\tau = 0.07$) & 1e-3 & Gaussian & \xmark & .73 &	.23	&.53$\pm$.13 &	2.21$\pm$.35 &	.29$\pm$.03 \\
        & SOM-SimCLR ($\tau=1$)  & 1e-3 & Gaussian & \xmark & .70	& .20 &	.48$\pm$.16	& 1.87$\pm$.30	&.50$\pm$.07
\\\hline
         & SOM-CPC (ours) & 
          1e-5 & Gaussian & \xmark & .78	&	.27	& 	.59$\pm$.11	&	1.03$\pm$.11	&	.04$\pm$.01

          \\
          &  & 
          1e-4 & Gaussian & \xmark & .78	&	.27	&	.61$\pm$.10	&	\textbf{1.01$\pm$.10}	&	.06$\pm$.02

          \\
          * & & 
          1e-3 & Gaussian & \xmark & .78	&	.27	&	.61$\pm$.12	&	1.02$\pm$.09	&	.03$\pm$.01

          \\
          & & 
          1e-2 & Gaussian & \xmark & \textbf{.79}	&	\textbf{.28}	&	.60$\pm$.11	&	1.08$\pm$.11	&	.19$\pm$.04

          \\
          & & 
          .1 & Gaussian & \xmark & \textbf{.79}	&	\textbf{.28}	&	\textbf{.65$\pm$.07}	&	1.09$\pm$.09	&	.19$\pm$.04
          \\\hdashline
           \hspace{-0.5cm}\parbox[t]{2mm}{\multirow{10}{*}{\rotatebox[origin=c]{90}{\underline{~~~~~~~~~~~~~Ablations~~~~~~~~~~~~~}} }} &   & 
          1e-3 & Gaussian & \cmark & .78	&	.27	&	.62$\pm$.10	&	1.02$\pm$.12	&	.07$\pm$.03

          \\
          &  & 
          1e-2 & Gaussian & \cmark & .78	&	.27	&	.60$\pm$.10	&	1.07$\pm$.09	&	.17$\pm$.04

          \\
          &  & 
          .1 & Gaussian & \cmark & .78	&	.27	&	.62$\pm$.10	&	1.06$\pm$.11	&	.08$\pm$.02

          \\
          & & 
          1 & Gaussian & \cmark & .78	&	.27	&	.59$\pm$.12	&	1.04$\pm$.12	&	.08$\pm$.02
          \\
          & & 
          1e-3 & Plus & \xmark & \textbf{.79}	&	\textbf{.28}	&	.61$\pm$.11	&	1.35$\pm$.24	&	.26$\pm$.08

          \\
          & & 
          1e-3 & Plus & \cmark & \textbf{.79}	&	\textbf{.28}	&	.60$\pm$.10	&	1.38$\pm$.28	&	.27$\pm$.08
        \\
        & SOM-CPC ($\tau=0.07$, $\operatorname{sim} =$ cosine sim.)& 1e-3 & Gaussian & \xmark & 
        .78 &	.27 &	.60$\pm$.12 &	1.36$\pm$.13 & .07$\pm$.02
        \\ 
        & SOM-CPC ($\tau=1$, $\operatorname{sim} =$ cosine sim.) & 1e-3 & Gaussian & \xmark & 
        .73	& .27	& .58$\pm$.11 &	1.43$\pm$.15 &	\textbf{.03$\pm$.01}
        \\
        & SOM-CPC ($\tau=0.07$, $\operatorname{sim} =$ dot prod.) & 1e-3 & Gaussian & \xmark & 
        \textbf{.79}	& \textbf{.28} &	.62$\pm$.10 &	1.06$\pm$.11 & .06$\pm$.02
        \\
    	& CPC + SOM (disjoint) & - & Gaussian & - & \textbf{.79}	&	\textbf{.28}	&	.62$\pm$.11	&	1.21$\pm$.11 & .52$\pm$.04
          \\\hline
    \end{tabular}
\end{table}

\clearpage
\subsection{Audio experiments}
\label{app:audio}

\subsubsection{Training details}
\label{app:audio_architectures}

Audio streams were encoded in windows of 0.01 seconds (=160 samples). For CPC, GRU-DESOM, and SOM-CPC the contextual information of $L=127$ previous windows was aggregated using a GRU, equivalent as proposed by \citet{Oord2019}. The InfoNCE objective for (SOM-)CPC was computed on top of the last context vector of the GRU. To test different settings for the GRU-desom model, we distinguished GRU-DESOM that decodes only the last window, given the last context vector, and GRU-DESOM that decodes the full sequence of 128 windows from the respective context vectors. 

Table \ref{tab:audio_model_details} provides the model details of the encoder, and the decoder for the DESOM benchmarks. The reconstruction loss of the DESOM model was found to be hampered in its optimization when using the encoder architecture, as adopted for the SOM-CPC model. As such, the downsampling factor of the encoder was reduced for the DESOM model to enable minimization of the task loss during training.


\begin{table*}[ht]
\tiny
\caption{Model details for the audio experiments in \cref{sec:audio}.}.
\centering
\label{tab:audio_model_details}
\begin{tabular}{llcccccc}
\textbf{Layer type} & \textbf{Output size} & \textbf{Channels} & \textbf{Activation} & \textbf{Kernel size} & \textbf{Strides} & \textbf{Dilation} &\textbf{Padding} \\
\hline\hline
\rowcolor{Gray}
\multicolumn{8}{l}{\textbf{\hspace{-0.2cm}Encoder for SOM-CPC and CPC}}\\\hline
Conv1D  & bs $\times \ 512 \times 32$ & 512 & ReLU  & $10$  & $5$ & $1$ & $3$ \\[0.1em]
Conv1D  & bs $\times \ 512 \times 8$ & 512 & ReLU  & $8$  & $4$ & $1$ & $2$ \\[0.1em]
Conv1D  & bs $\times \ 512 \times 4$ & 512 & ReLU  & $4$  & $2$ & $1$ & $1$ \\[0.1em]
Conv1D  & bs $\times \ 512 \times 2$ & 512 & ReLU  & $4$  & $2$ & $1$ & $1$ \\[0.1em]
Conv1D  & bs $\times \ 512$ & 512 & ReLU  & $4$  & $2$ & $1$ & $1$ \\[0.1em]
GRU &  bs $\times \ 512$ & 512 & - & - & - & - & - \\[0.1em]
\hline
\rowcolor{Gray}
\multicolumn{8}{l}{\textbf{\hspace{-0.2cm}Encoder for (GRU-)DESOM}}\\\hline
Conv1D  & bs $\times \ 512 \times 32$ & 512 & ReLU  & $10$  & $5$ & $1$ & $3$ \\[0.1em]
Conv1D  & bs $\times \ 512 \times 8$ & 512 & ReLU  & $8$  & $4$ & $1$ & $2$ \\[0.1em]
Conv1D  & bs $\times \ 512 \times 4$ & 512 & ReLU  & $4$  & $2$ & $1$ & $1$ \\[0.1em]
Conv1D  & bs $\times \ 512 \times 2$ & 512 & ReLU  & $4$  & $2$ & $1$ & $1$ \\[0.1em]
Conv1D  & bs $\times \ 512 \times 2$ & 512 & ReLU  & $4$  & $1$ & $1$ & same \\[0.1em]
Flatten & bs $\times 1024 $ & - & - & - & - & - & - \\[0.1em]
GRU &  bs $\times 1024$ & 1024 & - & - & - & - & - \\[0.1em]
\hline
\rowcolor{Gray}
\multicolumn{8}{l}{\textbf{\hspace{-0.2cm}Decoder for (GRU-)DESOM}}\\\hline
Unflatten & bs $\times \ 512 \times 2$ & - & - & - & - & - & - \\[0.1em]
Conv1D  & bs $\times \ 512 \times 2$ & 512 & ReLU  & $4$  & $1$ & $1$ & same\\[0.1em]
ConvTranspose1D  & bs $\times \ 512 \times 4$ & 512 & ReLU  & $4$  & $2$ & $1$ & $1$ \\[0.1em]
ConvTranspose1D  & bs $\times \ 512 \times 8$ & 512 & ReLU  & $4$  & $2$ & $1$ & $1$ \\[0.1em]
ConvTranspose1D  & bs $\times \ 512 \times 32$ & 512 & ReLU  & $8$  & $4$ & $1$ & $2$ \\[0.1em]
ConvTranspose1D  & bs $\times \ 1 \times 160$ & 512 & ReLU  & $10$  & $5$ & $1$ & $3$ (+ output pad = $1$) \\[0.1em]\hline
\end{tabular}
\end{table*}

For training of SOM-CPC, we followed the settings from \citet{Oord2019} and set $P=12$. The number of negative samples was set to $N = 10$, which were drawn randomly from the entire training set. All deep-SOM models were trained for maximally 3000 epochs, using the Adam optimizer \cite{kingma2014adam} with a learning rate of $1\text{e-}4$ and a batch size of 8. One epoch was defined as a push through of one sequence of $128$ windows (or 1 window for DESOM) from each recording. The best model was selected based on the lowest validation task loss. 

The supervised linear classifier and disjoint SOM training on top of the frozen CPC embeddings were trained with a batch size of 128, and a learning rate of $1\text{e-}4$ and $1\text{e-}2$, respectively. Both models were stopped upon convergence of the validation loss (which was after 200 and 250 epochs, respectively).

\subsubsection{Extended Results \label{app:extended_audio_results}}
Quantitative results of the audio experiments can be found in \cref{tab:audio_results}. For this application, Cohen's kappa is computed as the average over all data windows in the test set, which is different from the synthetic and physiological case, where it was computed as the average and one standard deviation across recordings. In these audio experiments, all windows from one recording contain the same speaker id label. Computing Cohen's kappa per-recording, i.e. with having the same label for all windows in that recording, is therefore inappropriate as the computation can not correct for correctness by chance. Table \ref{tab:audio_results} show that SOM-CPC clearly outperformed all variants of the (GRU-)DESOM model, and feature extraction using CPC, followed by PCA and linear or non-linear classification. 

Ablations with respect to the temperature $\tau$ and the similarity metric in the loss function indicated that simply changing the temperature value did hardly affect SOM-CPC performance. However, changing the similarity metric to be the cosine similarity caused a drop in clustering and classification performance, but only when using a temperature value of 1. This result is in line with the experimental findings in the physiological case (see \cref{app:extended_sleep_results}), and suggests that a low temperature value - which was found beneficial in the SimCLR objective \citep{Chen2020, Woo2022CoST:Forecasting} - did not have much influence on SOM-CPC performance when using the dot product as the similarity metric.

\begin{figure*}[t]
\centering     \includegraphics[width=0.5\linewidth,trim={0cm 6cm 9.5cm 0cm},clip,page=1]{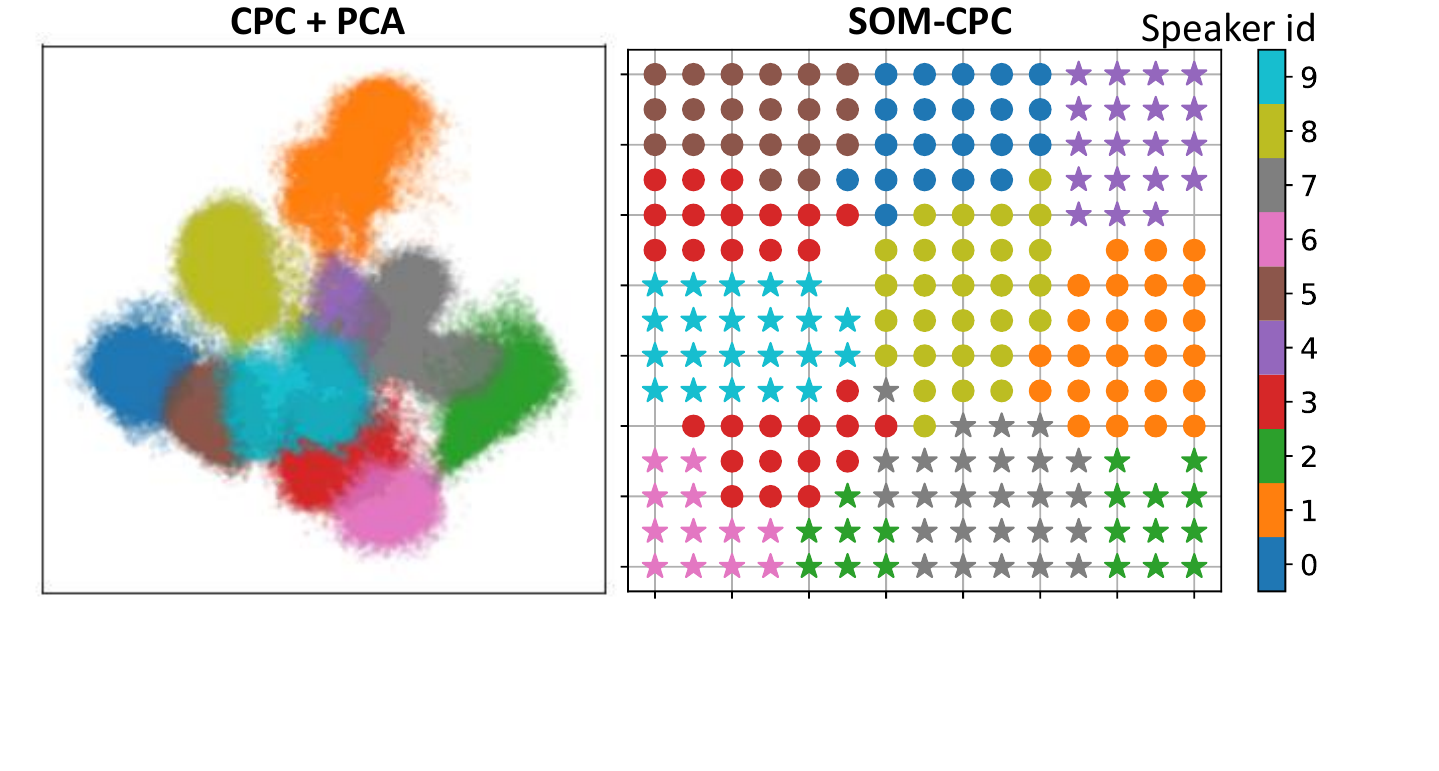}
    \vspace{-0.5cm}
    \caption{Projecting the test set on the 2D PCA space after CPC (with $F=128$) encoding, shows no division of the green and the red clusters in two sub-clusters, something that is visible in the SOM of the SOM-CPC model. These sub-clusters were found to relate to recordings that were made with different room acoustics.}
    \label{fig:audio_pca_vs_somcpc}
\end{figure*}

Figure \ref{fig:audio_pca_vs_somcpc} 
 compares the test set projection on the 2D PCA space, created on the CPC features (with $F=128$), to the SOM from the SOM-CPC model that is denoted with a $*$ in \cref{tab:audio_results} (and also visualized in \cref{fig:audio_som_maps}-right). This SOM reveals two separate clusters both for speaker 2 (green) and 3 (red), which were found to originate from recordings with different room acoustics (see \cref{sec:audio}). This division between recordings of the same speaker is not visible in the 2D PCA projection (\cref{fig:audio_pca_vs_somcpc}-left)


\begin{table}[t]
    \tiny
    \caption{Test set performance of various models trained on audio recordings. SOMs of models with a * are visualized in \cref{fig:audio_som_maps}. Bold values indicate the best performance per column (excluding the upper bound of the vanilla CPC model, which does not result in a 2D representation).}
    \label{tab:audio_results}
    \centering 
\begin{tabular}{p{0.1cm}lccc|llll}
        \hline
    	\rowcolor{Gray}
    	& \textbf{Model} & $\boldsymbol{\alpha}$ & $\mathcal{S}$ & $\mathcal{L}_{\text{SOM}} \operatorname{sg}[\cdot]$ &  \textbf{Purity} & \textbf{NMI} & \textbf{Cohen's kappa} & \textbf{TE}
    	\\\hline\hline
    	& CPC + linear classifier & - & - & - & - & - & 1.00 & - \\
    	& CPC + K-means & - & - & - & 1.00	&	.60	&	1.00 & -
    	\\\hline
    	& CPC ($F=2$)~+ linear classifier & - & - & - & - &	- &	~~.00 & -
    	\\
    	& CPC ($F=2$)~+ K-means & - & - & - & ~~.13	&	.01	&	~~.03 \\
    	& CPC + PCA + linear classifier & - & - & - & -	&	- &	~~.86 & -
    	\\
    	& CPC + PCA + K-means & - & - & - & ~~.89	&	.54	&	~~.88
 \\\hline
        & DESOM & 1e-5   & Gaussian & \xmark  & ~~.18	&	.03	&	~~.06	&	~~.14$\pm$.04
        \\
        & & 1e-4   & Gaussian & \xmark  & ~~.23	&	.08	&	~~.11	&	~~.34$\pm$.05
        \\
        & & 1e-3   & Gaussian & \xmark  & ~~.31	&	.13	&	~~.20	&	~~.68$\pm$.05
        \\
        & & 1e-2   & Gaussian & \xmark  & ~~.13	&	.00	&	~~.00	&	1.00$\pm$.00
        \\\hline
        & GRU-DESOM (reconstructing last window) & 1e-5   & Gaussian & \xmark  & ~~.19	&	.04	&	~~.08	&	\textbf{~~.13$\pm$.03}
        \\
        & & 1e-4   & Gaussian & \xmark  & ~~.26	&	.09	&	~~.15	&	~~.20$\pm$.04
        \\
        & & 1e-3   & Gaussian & \xmark  & ~~.31	&	.13	&	~~.21	&	~~.42$\pm$.08
        \\
        & & 1e-2   & Gaussian & \xmark  & ~~.32	&	.14	&	~~.22	&	~~.46$\pm$.06
        \\\hline
        & GRU-DESOM (reconstructing full sequence) & 1e-5   & Gaussian & \xmark  & ~~.19	&	.05	&	~~.08	&	~~.34$\pm$.05
        \\
        & & 1e-4   & Gaussian & \xmark  & ~~.30	&	.12	&	~~.20	&	~~.59$\pm$.07
        \\
        * & & 1e-3   & Gaussian & \xmark  & ~~.33	&	.14	&	~~.22	&	~~.78$\pm$.05
        \\
        & & 1e-2   & Gaussian & \xmark  & ~~.29	&	.12	&	~~.19	&	~~.57$\pm$.07
         \\\hline
        & SOM-CPC (ours) & 1e-5   & Gaussian & \xmark  & ~~.99	&	\textbf{.73}	&	~~.99	&	~~.14$\pm$.08
        \\
        & & 1e-4   & Gaussian & \xmark  & \textbf{1.00} &	.63	&	\textbf{1.00}	&	~~.24$\pm$.12
        \\
        * & & 1e-3   & Gaussian & \xmark  & \textbf{1.00}	&	.61	&	\textbf{1.00}	&	~~.33$\pm$.10
        \\
        & & 1e-2   & Gaussian & \xmark  & \textbf{1.00}	&	.61	&	~~.99	&	~~.33$\pm$.10
        \\\hdashline
       \hspace{-0.5cm}\parbox[t]{2mm}{\multirow{8}{*}{\rotatebox[origin=c]{90}{\underline{~~~~~~~~Ablations~~~~~~~~}} }}  & & 1e-3   & Gaussian & \cmark  & \textbf{1.00}	&	.61	&	~~.99	&	~~.28$\pm$.08
        \\
        & & 1e-2   & Gaussian & \cmark  & \textbf{1.00}	&	.61	&	~~.99	&	~~.35$\pm$.10
        \\
        & & .1   & Gaussian & \cmark  & \textbf{1.00}	&	.61	&	\textbf{1.00}	&	~~.35$\pm$.10
        \\
        & & 1   & Gaussian & \cmark  & \textbf{1.00	}&	.61	&	\textbf{1.00}	&	~~.38$\pm$.11
        \\ 
        & SOM-CPC ($\tau=0.07$, $\operatorname{sim} =$ cosine sim.)& 1e-3 & Gaussian & \xmark & 
        ~~.99 	&	.61	&	~~.99		& ~~.42$\pm$.12
        \\
        & SOM-CPC ($\tau=1$, $\operatorname{sim} =$ cosine sim.) & 1e-3 & Gaussian & \xmark & 
        ~~.88 	&	.55	&	~~.86	&	~~.17$\pm$.06
        \\
        & SOM-CPC ($\tau=0.07$, $\operatorname{sim} =$ dot prod.) & 1e-3 & Gaussian & \xmark & 
        \textbf{1.00} 	&	.61	&	~~.99	&	~~.38$\pm$.10
        \\
    	& CPC + SOM (disjoint) & - & Gaussian & - & \textbf{1.00}	&	.62	&	\textbf{1.00} & ~~.28$\pm$.11
        \\\hline
    \end{tabular}
\end{table}

\end{document}